\documentclass[10pt, journal,twocolumn]{IEEEtran}
\usepackage{amsmath,epsfig}
\usepackage{float}

\usepackage{tabularx}
\usepackage{graphicx}
\usepackage{epstopdf}
\usepackage{color}
\usepackage{subfigure}
\usepackage{multitoc}
\usepackage[font=footnotesize, labelfont=footnotesize]{caption}

\makeatletter
\newcommand*{\rom}[1]{\expandafter\@slowromancap\romannumeral #1@}
\makeatother

\begin{document}

\title{Structured Priors for Sparse-Representation-Based Hyperspectral Image Classification}

\author{Xiaoxia Sun,  Qing Qu, Nasser M. Nasrabadi,~\IEEEmembership{Fellow,~IEEE,} and Trac D. Tran,~\IEEEmembership{Senior Member,~IEEE}
\thanks{X.Sun, Q. Qu and T. D. Tran are with the Department of Electrical and Computer Engineering, The Johns Hopkins University, Baltimore, MD 21218 USA (e-mail: xsun9@jhu.edu; qqu2@jhu.edu; trac@jhu.edu). This work has been partially supported by NSF under Grants CCF-1117545, ARO under Grants 60219-MA, and ONR under grant N000141210765.} 
\thanks{N. M. Nasrabadi is with U.S. Army Research Laboratory, Adelphi, MD 20783
USA (e-mail: nnasraba@arl.army.mil).} }
%


\markboth{IEEE Geoscience and Remote Sensing Letter}
{Structured Priors for Sparse Representation Based Hyperspectral Image Classification}

%


\maketitle

\begin{abstract}
Pixel-wise classification, where each pixel is assigned to a predefined class, is one of the most important procedures in hyperspectral image (HSI) analysis.  By representing a test pixel as a linear combination of a small subset of labeled pixels, a sparse representation classifier (SRC) gives rather plausible results compared with that of traditional classifiers such as the support vector machine (SVM). Recently, by incorporating additional structured sparsity priors, the second generation SRCs have appeared in the literature and are reported to  further improve the performance of HSI. These priors are based on exploiting the spatial dependencies between the neighboring pixels, the inherent structure of the dictionary, or both. In this paper, we review and compare several structured priors for sparse-representation-based HSI classification. We also propose a new structured prior called the low rank group prior, which can be considered as a modification of the low rank prior. Furthermore, we will investigate how different structured priors improve the result for the HSI classification. 
\end{abstract}
\section{Introduction}
\label{sec:intro}

\PARstart{o}{ne} of the most important procedures in HSI is image classification, where the  pixels  are labeled to one of the classes based on their spectral characteristics. Due to the numerous demands in  mineralogy, agriculture and surveillance, the HSI classification task is developing very rapidly and a large number of techniques have been proposed to tackle this problem \cite{Plaza}.  Comparing with previous approaches, SVM is found highly effective on both computational efficiency and classification results.    A wide variety of SVM's modifications have been proposed to improve its performance. Some of them incorporate the contextual information in the classifiers \cite{GCampsValls,Chova}. Others design sparse SVM in order to pursue a sparse decision rule by using $\ell_1$-norm as the regularizer \cite{JZhu}.  

Recently, SRC has been proposed to solve many computer vision tasks \cite{JWright2,Marial}, where the use of sparsity as a prior often leads to state-of-the-art performance. SRC has also been applied to HSI classification \cite{Chen}, relying on the observation that hyperspectral pixels belonging to the same class approximately lie in the same low-dimensional subspace. In order to alleviate the problem introduced by the lack of sufficient training data, Haq \emph{et al}. \cite{Haq} proposed the homotopy-based SRC. Another way to solve the problem of insufficient training data is to employ the contextual information of neighboring pixels in the classifier, such as spectral-spatial constraint classification \cite{RJi}.

In SRC, a test sample $\mathbf{y}\in \mathbf{R}^P$, where $P$ is the number of spectral bands, can be written as a sparse linear combination of all the training pixels (atoms in a dictionary) as
\begin{equation}
\label{equ::l1recov}
\mathbf{\hat{x}} = \min_{\mathbf{x}}{\frac{1}{2}\lVert\mathbf{y-Ax}\rVert_2^2 + \lambda\lVert\mathbf{x}\rVert_1} ,
\end{equation}
where  $\mathbf{x}\in \mathbf{R}^N$, $\lVert\mathbf{x}\rVert_1 = \sum\limits_{i=1}^N |x_i |$  is $\ell_1$-norm. $\mathbf{A}=[\mathbf{a}_1,\mathbf{a}_2, \cdots, \mathbf{a}_N]$ is a structured dictionary formed from concatenation of several class-wise sub-dictionaries, $\{\mathbf{a}_i\}_{i=1,\dots, N}$ are the columns of $\mathbf{A}$ and $N$ is the total number of training samples from all the $K$ classes, and $\lambda$ is a scalar regularization parameter.

The class label for the test pixel $\mathbf{y}$ is determined by the minimum residual between $\mathbf{y}$ and its approximation from each class-wise sub-dictionary:
\begin{equation}
class(\mathbf{y}) = \arg\min_g{\lVert\mathbf{y} - \mathbf{A}\delta_g(\mathbf{x})\rVert_2^2},
\end{equation}
where $g\subset \{1, 2,\cdots, K\}$  is the group or class index, and $\delta_g(\mathbf{x})$ is the indicator operation zeroing out all elements of $\mathbf{x}$ that do not belong to the class $g$.

  In  the case of HSI, SRC always suffers from the non-uniqueness or instability of the sparse coefficients due to the high mutual coherency of the dictionary \cite{ MIordache}. Fortunately, a better reconstructed signal and a more robust representation can be obtained by either exploring the dependencies of neighboring pixels or exploiting the inherent dictionary structure.   Recently, structured priors have been incorporated  into  HSI classification \cite{Chen}, which can be sorted into three categories. ({\it a}) Priors that only exploit the correlations and dependencies among the neighboring spectral pixels or their sparse coefficient vectors, which includes joint sparsity \cite{Ewout}, graph regularized Lasso (referred as the Laplacian regularized Lasso) \cite{Gao} and  the low-rank Lasso \cite{Liu}.   ({\it b}) Priors that only exploit the inherent structure of the dictionary,  such as group Lasso \cite{Rakotomamonjy}.  ({\it c}) Priors that enforce structural information on both sparse coefficients and dictionary,  such as collaborative group Lasso \cite{Kim} and  collaborative hierarchical Lasso (CHiLasso) \cite{Sprechmann}. Besides SRC, structured sparsity prior can also be incorporated into other classifiers such as the logistic regression classifiers \cite{Qian}.  

 The main contributions of this paper are \textsl{(a)} to assess the SRC performance using various structured sparsity priors for HSI classification, and \textsl{(b)} to propose a conceptually similar prior to CHiLasso, which is called the low-rank group prior. This prior is based on the assumption that pure or mixed pixels from the same classes are highly correlated and can be represented by a combination of sparse low-rank groups (classes).   The proposed prior takes advantage of both the group sparsity prior, which enforces sparsity across the groups, and the low rank prior, which encourages sparsity within the groups, by only using  one regularizer.  

In the following sections, we investigate the roles of different structured priors imposed on the SRC optimization algorithm. Starting with the classical sparsity  $\ell_1$-norm prior, we then introduce several different priors with experimental results. The structured priors discussed are joint sparsity, Laplacian sparsity, group sparsity, sparse group sparsity,  low-rank  and low-rank  group  prior.

\section{HSI Classification Via Different Structured Sparse Priors}

\label{sec:pagestyle}
\subsection{Joint Sparsity Prior}
In HSI, pixels within a small neighborhood usually consist of similar materials. Thus, their spectral characteristics are highly correlated. The spatial correlation between neighboring pixels can be indirectly incorporated through a joint sparsity model (JSM) \cite{Tropp} by assuming that the underlying sparse vectors associated with these pixels share a common sparsity support. 
Consider pixels in a small neighborhood of $T$ pixels. Let $\mathbf{Y}\in \mathbf{R}^{P\times T}$ represent a matrix whose columns correspond to pixels in a spatial neighborhood in a hyperspectral image. Columns of $\mathbf{Y} =[ \mathbf{y}_1,\mathbf{y}_2, \cdots, \mathbf{y}_T]$ can be represented as a linear combination of dictionary atoms $\mathbf{Y} = \mathbf{AX}$, where $\mathbf{X}=[\mathbf{x}_1, \mathbf{x}_2, \cdots, \mathbf{x}_T]\in \mathbf{R}^{  N \times T}$ represents a sparse matrix.
In JSM, the sparse vectors of $T$ neighboring pixels, which are represented by the $T$ columns of $\mathbf{X}$, share the same support.  Therefore, $\mathbf{X}$ is a sparse matrix with only few nonzero rows. The row-sparse matrix X can be recovered by solving the following Lasso problem
\begin{equation}
\min_{\mathbf{X}}{\frac{1}{2}\lVert\mathbf{Y-AX}\rVert_F^2 + \lambda\lVert\mathbf{X}\rVert_{1,2}} ,
\end{equation}
where $\lVert \mathbf{X} \rVert_{1,2}=\sum\limits_{i=1}^N\lVert \mathbf{x}^i \rVert_2$ is an $\ell_{1,2}$-norm and $\mathbf{x}^i$ represents the $i$th row of $\mathbf{X}$.

The label for the center pixel $\mathbf{y}_c$  is then determined by the minimum total residual error
\begin{equation}
class(\mathbf{y}_c) = \arg\min_g{\lVert\mathbf{Y} - \mathbf{A}\delta_g(\mathbf{X})\rVert_F^2},
\end{equation}
where $\delta_g(\mathbf{X})$ is the indicator operation zeroing out all  the elements of $\mathbf{X}$ that do not belong to the class $g$.

\subsection{Laplacian Sparsity Prior}
In sparse representation, due to the high coherency of the dictionary atoms, the recovered sparse coefficient vectors for multiple neighboring pixels could be partially different even when the neighboring pixels are highly correlated, and this may led to misclassification. As mentioned in the previous section, joint sparsity is able to solve such a problem by enforcing multiple pixels to select exactly the same atoms. However, in many cases, when the neighboring pixels fall on the boundary between several homogeneous regions, the neighboring pixels will belong to several distinct classes (groups) and should use different sets of sub-dictionary atoms. Laplacian sparsity enhances the differences between sparse coefficient vectors of the neighboring pixels that belong to different clusters. We introduce the weighting matrix $\mathbf{W}$, where $\mathbf{w}_{ij}$ characterizes the similarity between a pair of pixels $y_i$ and $y_j$ within a neighborhood. Optimization with an additional Laplacian sparsity prior can be expressed as
\begin{equation}
\label{eq:lap_1}
\min_{\mathbf{X}}{\frac{1}{2}\lVert\mathbf{Y-AX}\rVert_F^2 + \lambda_1\lVert\mathbf{X}\rVert_1 + \lambda_2\sum\limits_{i,j}w_{ij}\lVert\mathbf{x}_i-\mathbf{x}_j\rVert^2_2},
\end{equation}
where $\lambda_1$ and $\lambda_2$ are the regularization parameters. The matrix $\mathbf{W}$ is used to characterize the similarity among neighboring pixels in the spectra space. Similar pixels will possess larger weights, and therefore, enforcing the differences between the corresponding sparse coefficient vectors to become smaller, and similarly allowing the difference between sparse coefficient vectors of dissimilar pixels to become larger. Therefore, the Laplacian sparsity prior is more flexible than the joint sparsity prior in that it does not always force all the neighboring pixels to have the same common support. In this paper, the weighting matrix is computed using the sparse subspace clustering method in \cite{Elhamifar}. Note that this grouping constraint is enforced on the testing pixels instead of the dictionary atoms, which is different from group sparsity. Let $\mathbf{L=I-D}^{-1/2}\mathbf{WD}^{-1/2}$ be the normalized symmetric Laplacian matrix \cite{Elhamifar}, where $\mathbf{D}$ is the degree matrix computed from $\mathbf{W}$. We can rewrite the equation (\ref{eq:lap_1}) as
\begin{equation}
\min_{\mathbf{X}}{\frac{1}{2}\lVert\mathbf{Y-AX}\rVert_F^2 + \lambda_1\lVert\mathbf{X}\rVert_1 + \lambda_2tr(\mathbf{XLX}^T)}.
\end{equation}
The above equation can be solved by a modified feature-sign search algorithm \cite{Gao}. 

\subsection{Group Sparsity Prior}
The SRC dictionary has an inherent group-structured property since it is composed of several class sub-dictionaries, i.e., the atoms belonging to the same class are grouped together to form a sub-dictionary. In sparse representation, we classify pixels by measuring how well the pixels are represented by each sub-dictionary. Therefore, it would be reasonable to enforce the pixels to be represented by groups of atoms instead of individual ones. This could be accomplished by encouraging coefficients of only certain groups to be active and the remaining groups inactive. Group Lasso \cite{Rakotomamonjy}, for example, uses a sparsity prior that sums up the Euclidean norm of every group coefficients. It will dominate the classification performance especially when the input pixels are inherently mixed pixels. Group Lasso optimization can be represented as 
\begin{equation}
\label{eq:gs}
\min_{\mathbf{x}}{\frac{1}{2}\lVert\mathbf{y-Ax}\rVert_2^2 + \lambda\sum\limits_{g\in G}w_g\lVert\mathbf{x}_g\rVert_2},
\end{equation}
where $g \subset \{ G_1, G_2, \cdots, G_K\}$, $\sum\limits_{g\in G}\lVert\mathbf{x}_g\rVert_2$ represents the group sparse prior defined in terms of $K$ groups, $w_g$ is the weight and is usually set to the square root of the cardinality of the corresponding group to compensate for the different group sizes. Here, $\mathbf{x}_g$ refers to the coefficients of each group. The above group sparsity can be easily extended to the case of multiple neighboring pixels by extending problem (\ref{eq:gs}) to collaborative group Lasso, which is formulated as 
\begin{equation}
\min_{\mathbf{X}}{\frac{1}{2}\lVert\mathbf{Y-AX}\rVert_F^2 + \lambda\sum\limits_{g\in G}w_g\lVert\mathbf{X}_g\rVert_2},
\label{eq:cgs}
\end{equation}
where $\sum\limits_{g\in G}\lVert\mathbf{X}_g\rVert_2$ represents a collaborative group Lasso regularizer defined in terms of  group and $\mathbf{X}_g$ refers to each of the group coefficient matrix. When the group size is reduced to one, the group Lasso degenerates into a joint sparsity Lasso.

\subsection{Sparse Group Sparsity Prior}

In the formulations (\ref{eq:gs}) and (\ref{eq:cgs}), the coefficients within each group are not sparse, and all the atoms in the selected groups could be active.  If the sub-dictionary is overcomplete, then it is necessary to enforce sparsity within each group. To achieve sparsity within the groups, an $\ell_1$-norm regularizer  can be  added to the group Lasso (\ref{eq:gs}), which can be written as
\begin{equation}
\label{eq:sgs}
\min_{\mathbf{x}}{\frac{1}{2}\lVert\mathbf{y-Ax}\rVert_2^2 + \lambda_1\sum\limits_{g\in G}w_g\lVert\mathbf{x}_g\rVert_2 +\lambda_2\lVert\mathbf{x}\rVert_1}.
\end{equation}

Similarly, Eq. (\ref{eq:sgs}) can be easily extended to the multiple feature case, which can be written as

\begin{equation}
\label{eq:chilasso}
\min_{\mathbf{X}}{\frac{1}{2}\lVert\mathbf{Y-AX}\rVert_F^2 + \lambda_1\sum\limits_{g\in G}w_g\lVert\mathbf{X}_g\rVert_2+ \lambda_2\sum\limits_{g\in G}w_g\lVert\mathbf{X}_g\rVert_1}.
\end{equation}

Optimization problem (\ref{eq:sgs}) is referred in the literature as the sparse group Lasso  and problem (\ref{eq:chilasso}) as the collaborative hierarchical Lasso (CHiLasso) \cite{Sprechmann}.

\subsection{Low Rank/Group Sparsity Prior}

 Based on the fact that spectra of neighboring pixels are highly correlated, it is reasonable to enforce the low rank sparsity prior on their coefficient matrix. The low rank prior is more flexible when compared with the joint sparsity prior which strictly enforces the row sparsity. Therefore, when neighboring pixels are composed of small non-homogeneous regions, the low rank sparsity prior outperforms the joint sparsity prior. Low rank sparse recovery problem has been well studied in  \cite{Liu} and is stated as the following Lasso problem
\begin{equation}
\min_{\mathbf{X}}{\frac{1}{2}\lVert\mathbf{Y-AX}\rVert_F^2 + \lambda\lVert\mathbf{X}\rVert_{*}} ,
\end{equation}
where $\lVert\mathbf{X}\rVert_{*}$ is the nuclear norm \cite{Liu}.

\begin{figure}[ht]
\centering
\subfigure[]{

\label{Fig.sub.2}
\centering
\includegraphics[width=0.1\textwidth]{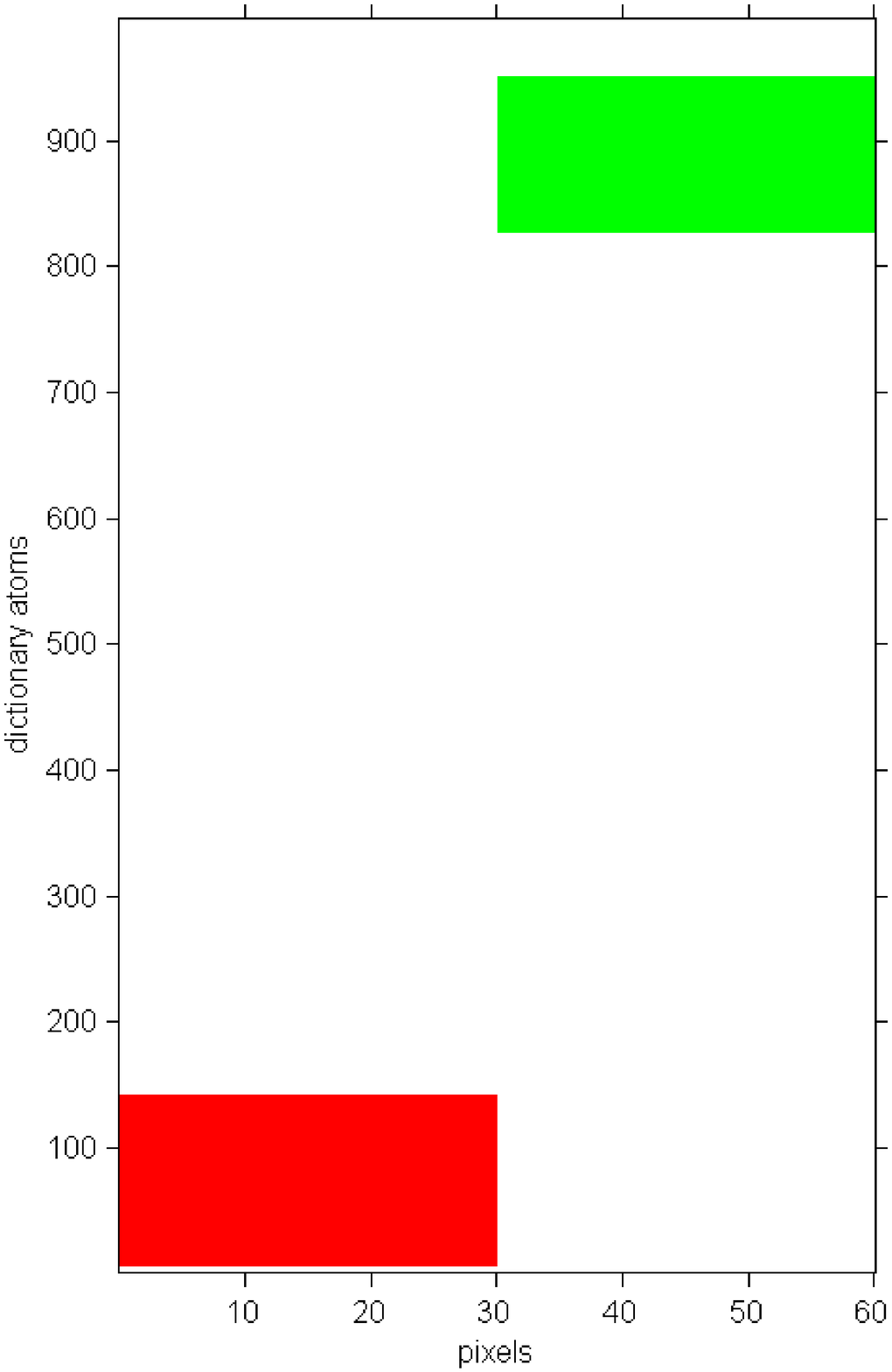}}
\subfigure[]{
\label{Fig.sub.1}
\includegraphics[width=0.1\textwidth]{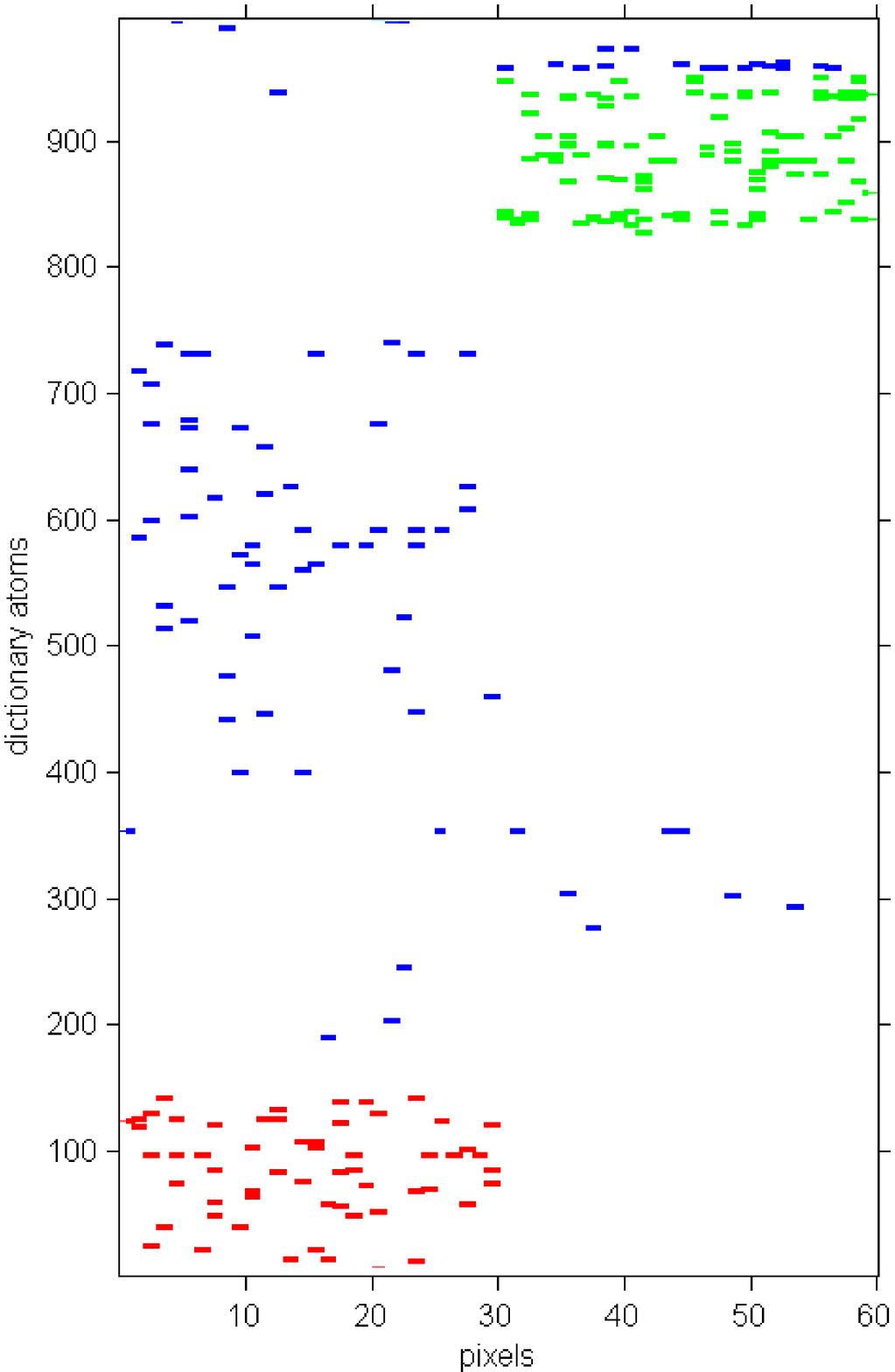}}
\subfigure[]{
\label{Fig.sub.2}
\includegraphics[width=0.1\textwidth]{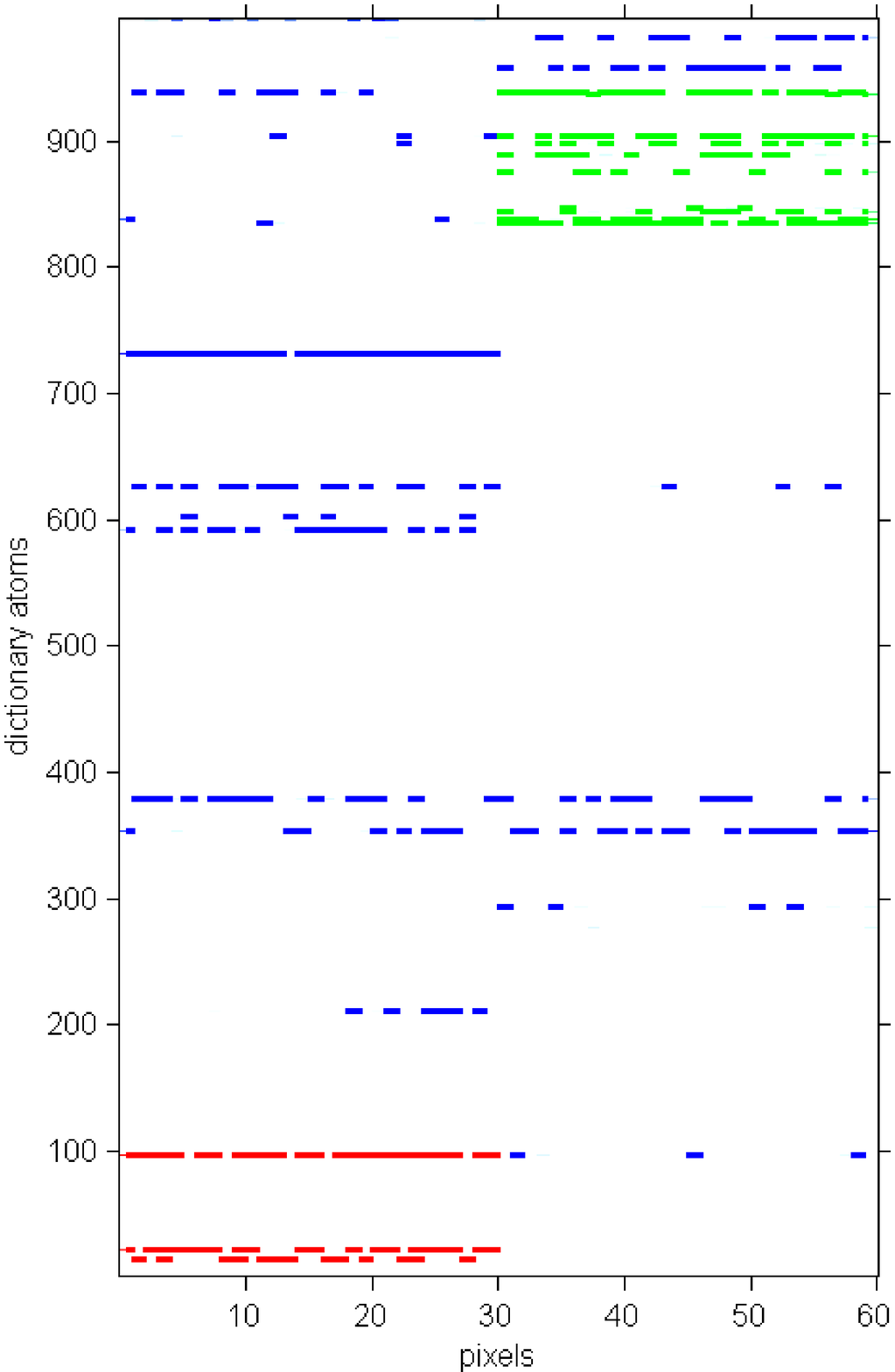}}
\subfigure[]{
\label{Fig.sub.2}
\includegraphics[width=0.1\textwidth]{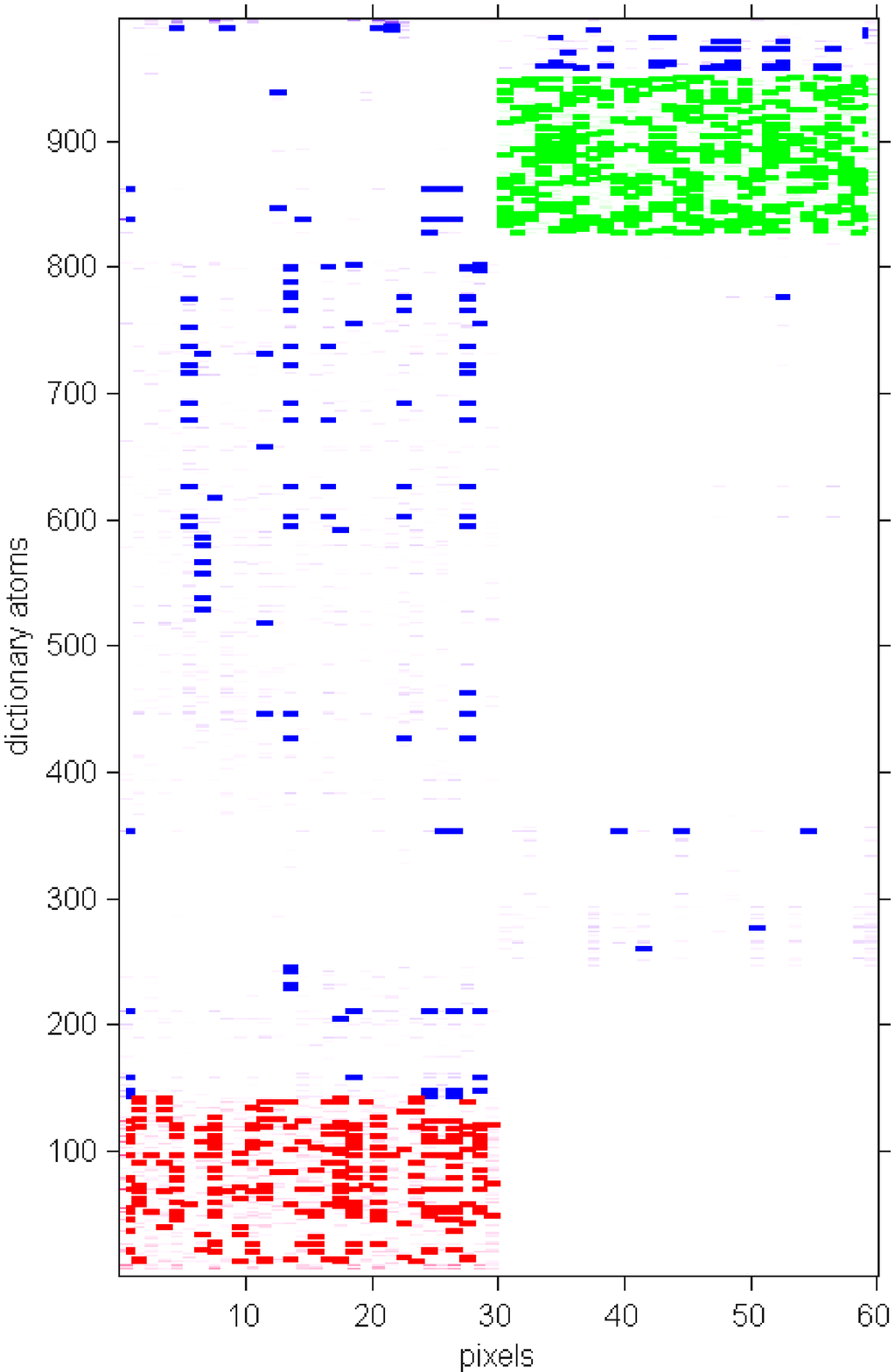}}
\subfigure[]{
\label{Fig.sub.2}
\includegraphics[width=0.1\textwidth]{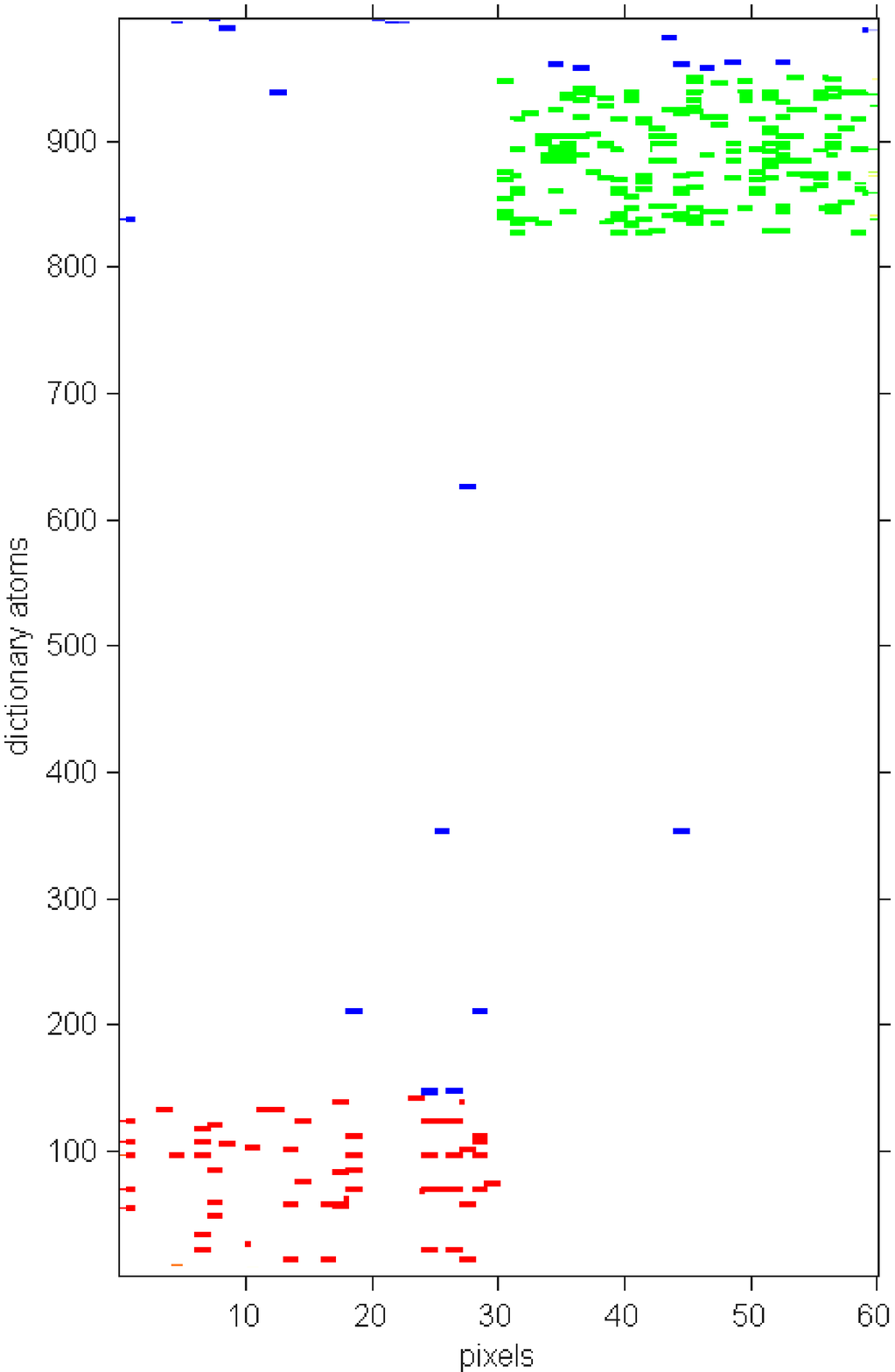}}
\subfigure[]{
\label{Fig.sub.2}
\includegraphics[width=0.1\textwidth]{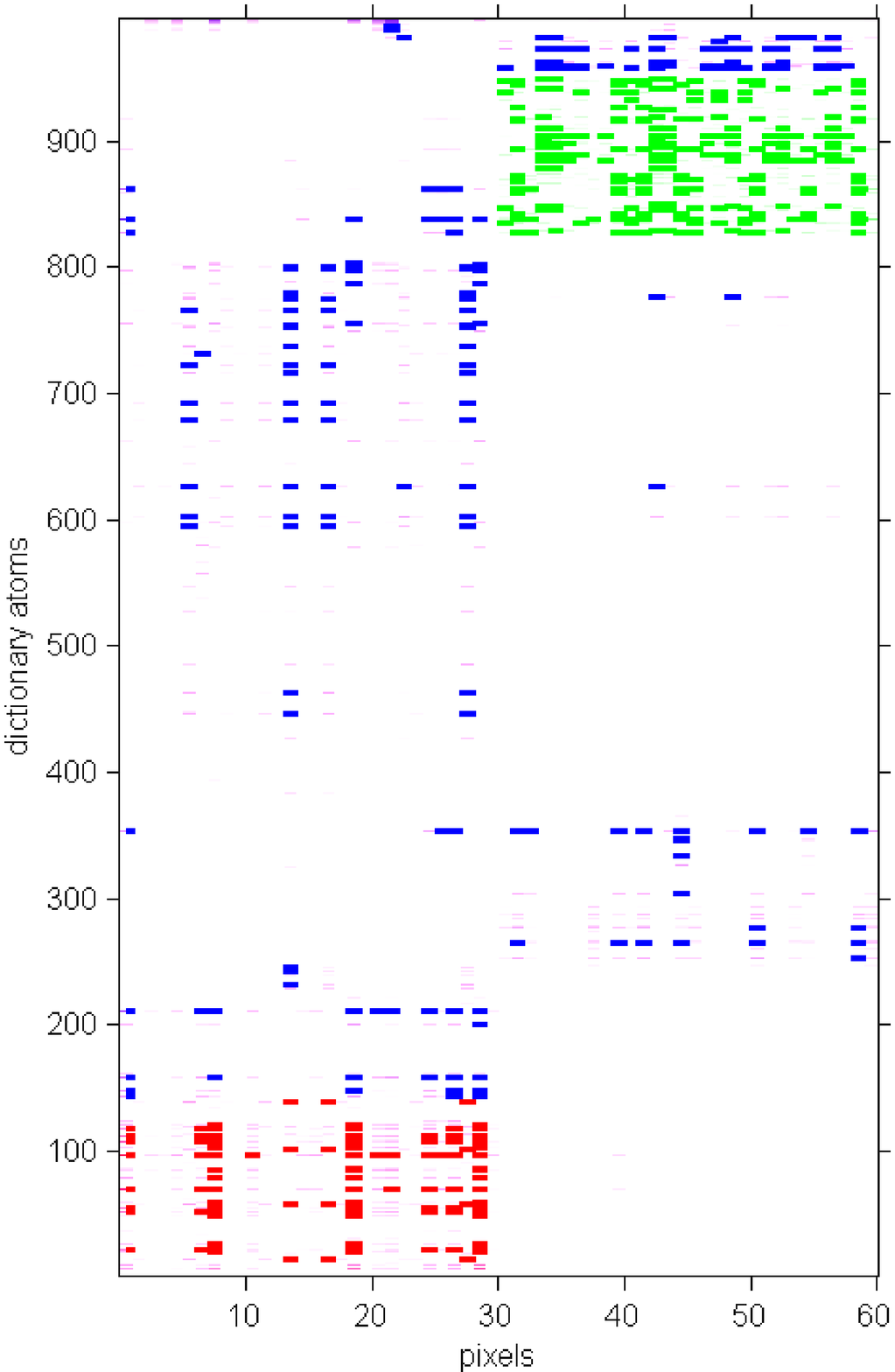}}
\subfigure[]{
\label{Fig.sub.2}
\includegraphics[width=0.1\textwidth]{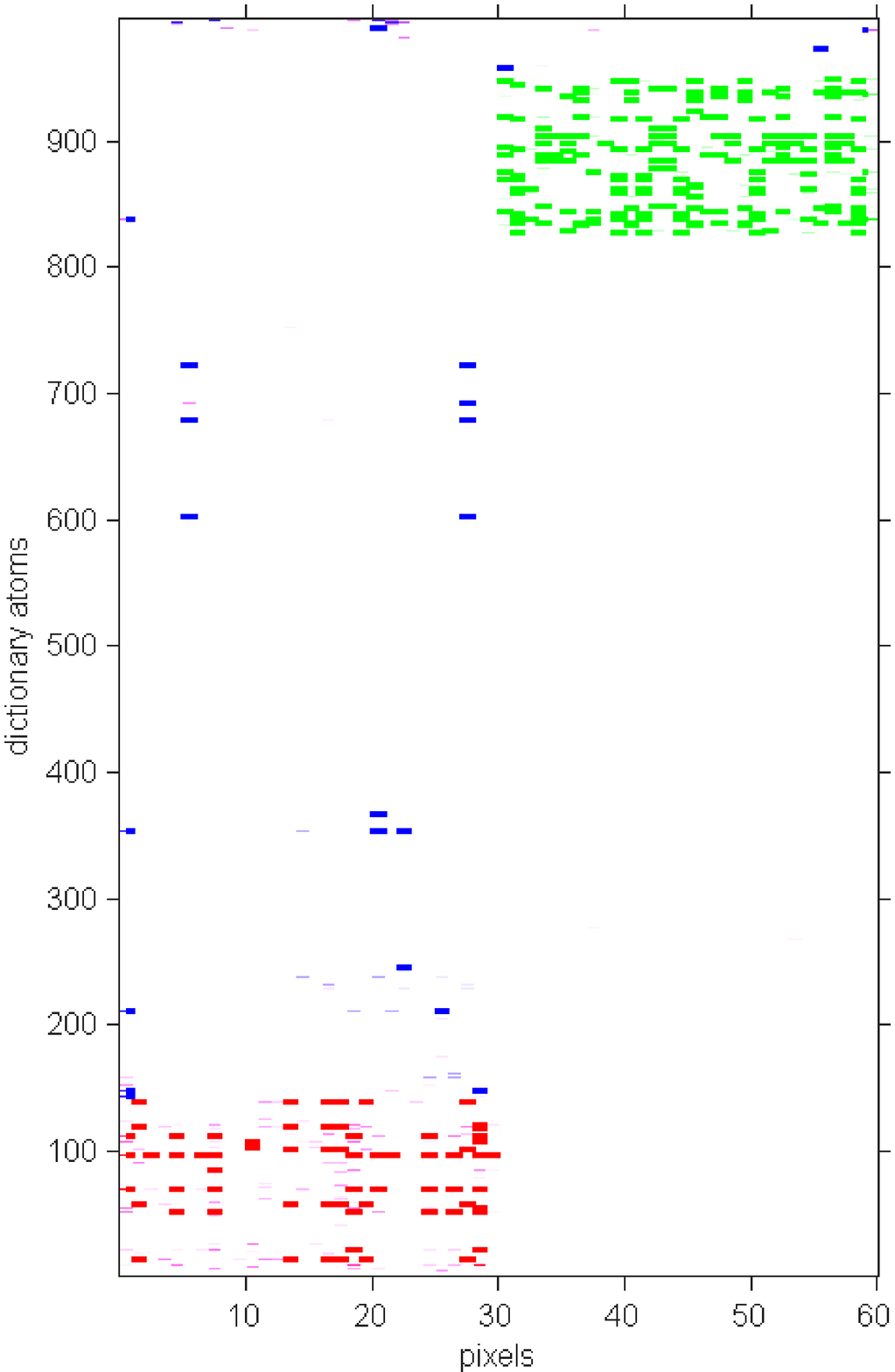}}
\subfigure[]{
\label{Fig.sub.2}
\includegraphics[width=0.1\textwidth]{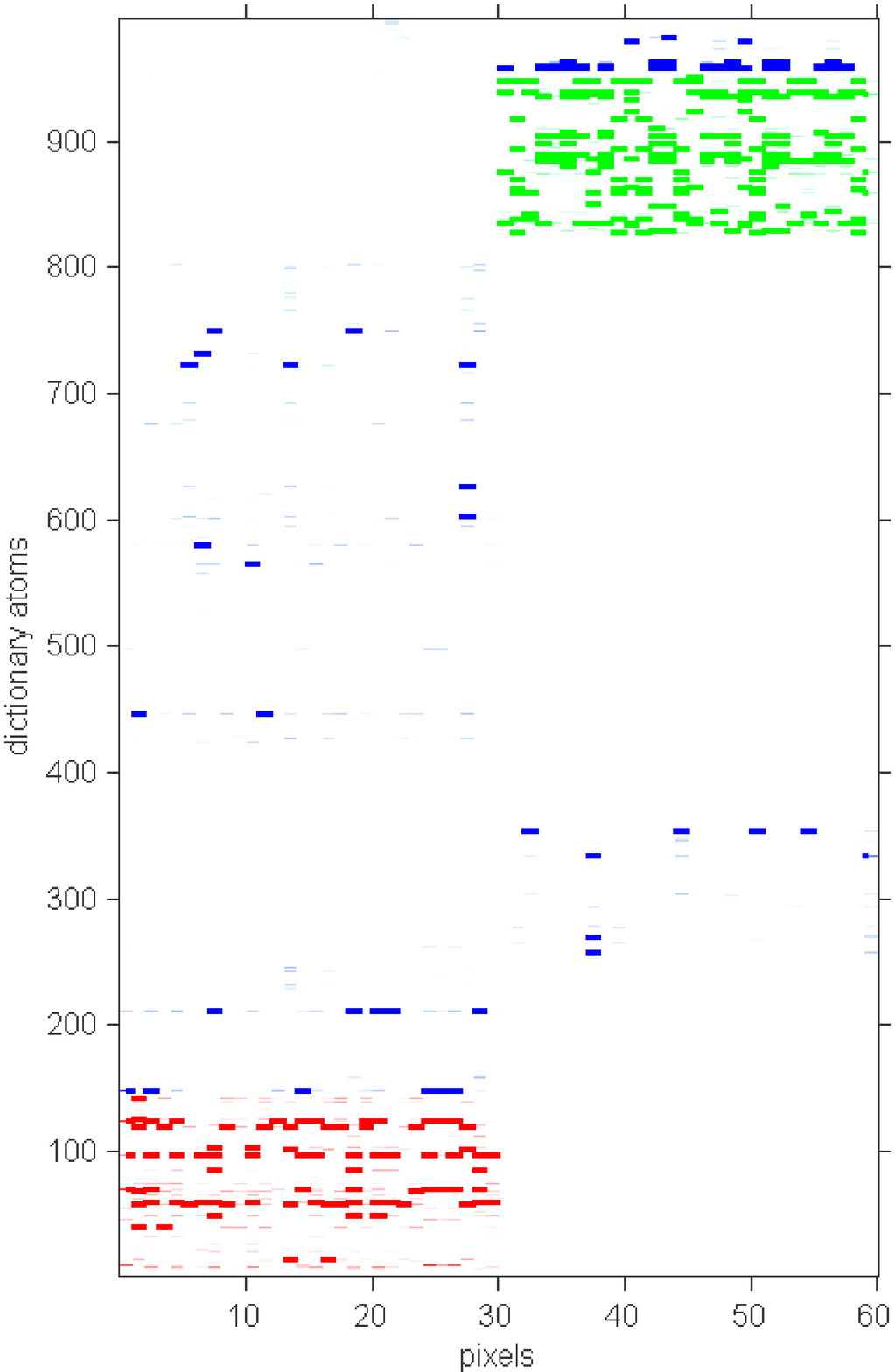}}

\caption{Sparsity patterns for the toy example: (a) desired sparsity regions, (b) $\ell_1$ minimization using ADMM, (c) joint sparsity,  (d)   collaborative group sparsity, (e)  collaborative sparse group sparsity, (f)  low rank sparsity, (g)  low rank group sparsity and (h) Laplacian sparsity via FFS.}  
\label{Fig.lable}
\label{fig:sp}
\end{figure}

\begin{figure*}[ht]
\centering
\begin{minipage}[b]{0.12\linewidth}
\includegraphics[width=\textwidth]{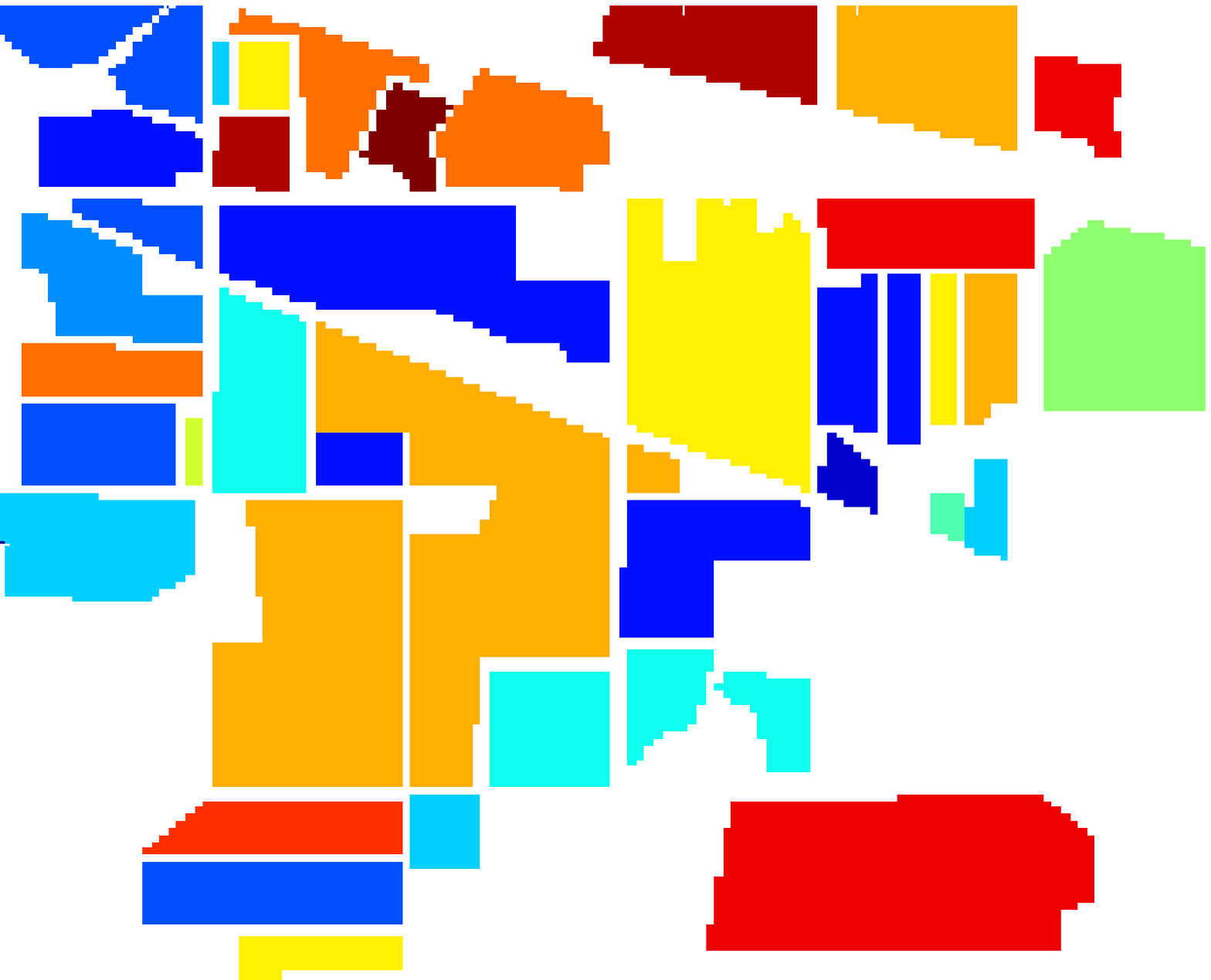}
\caption*{(a)}
\label{fig:figure1}
\end{minipage}
\hspace{0.2cm}
\begin{minipage}[b]{0.12\linewidth}
\centering
\includegraphics[width=\textwidth]{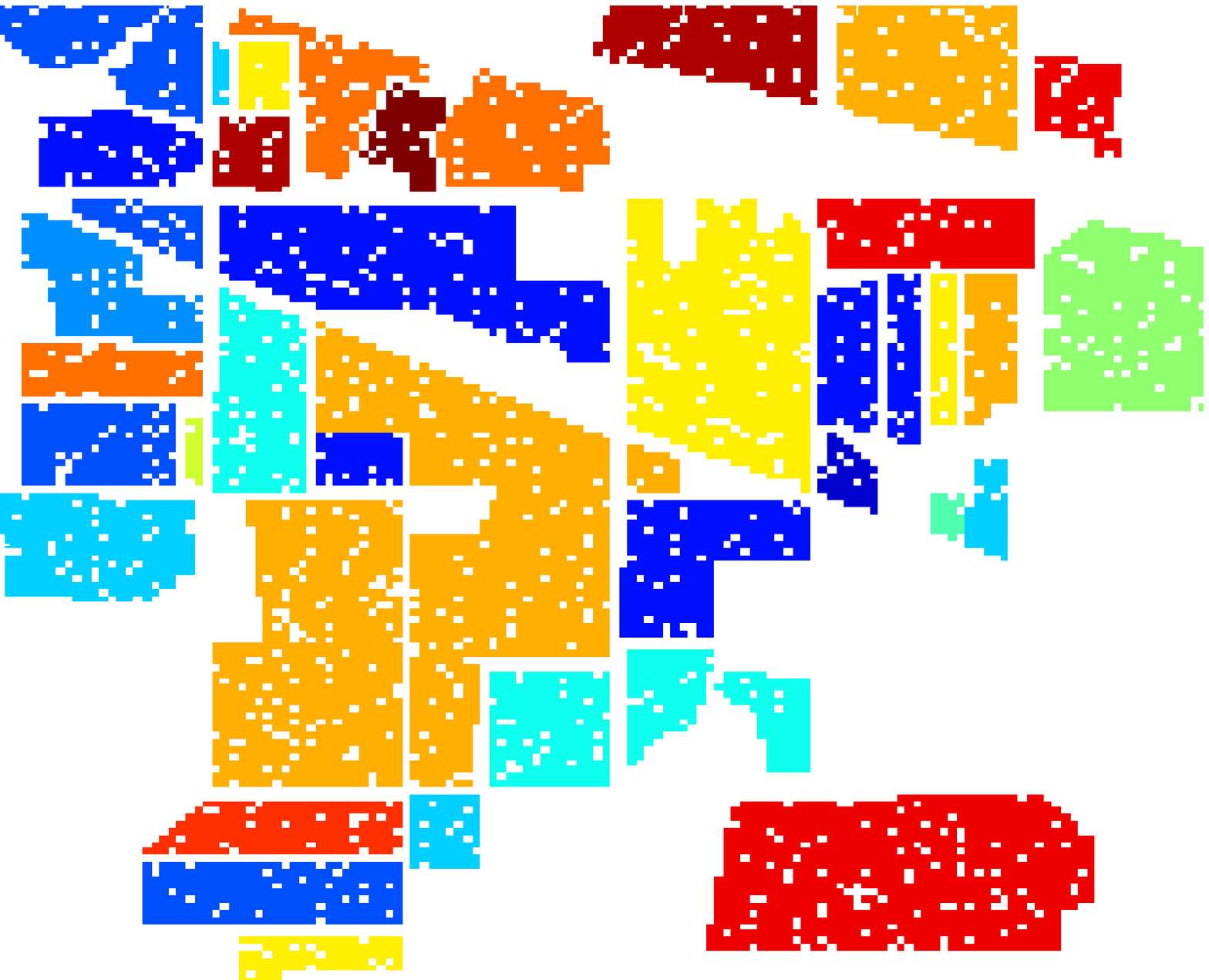}
\caption*{(b)}
\label{fig:figure2}
\end{minipage}
\hspace{0.2cm}
\begin{minipage}[b]{0.12\linewidth}
\centering
\includegraphics[width=\textwidth]{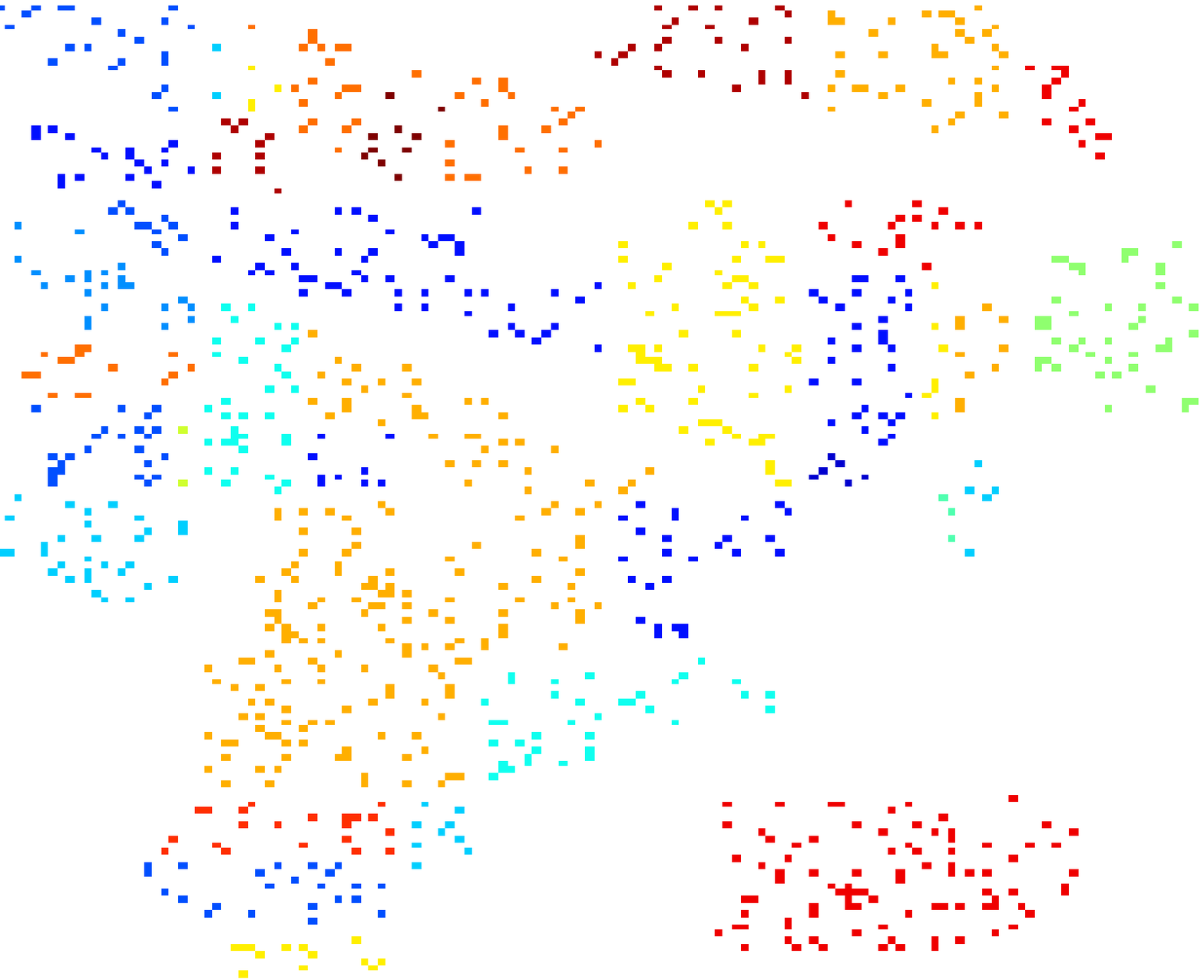}
\caption*{(c)}
\label{fig:figure2}
\end{minipage}
\hspace{0.2cm}
\begin{minipage}[b]{0.12\linewidth}
\centering
\includegraphics[width=\textwidth]{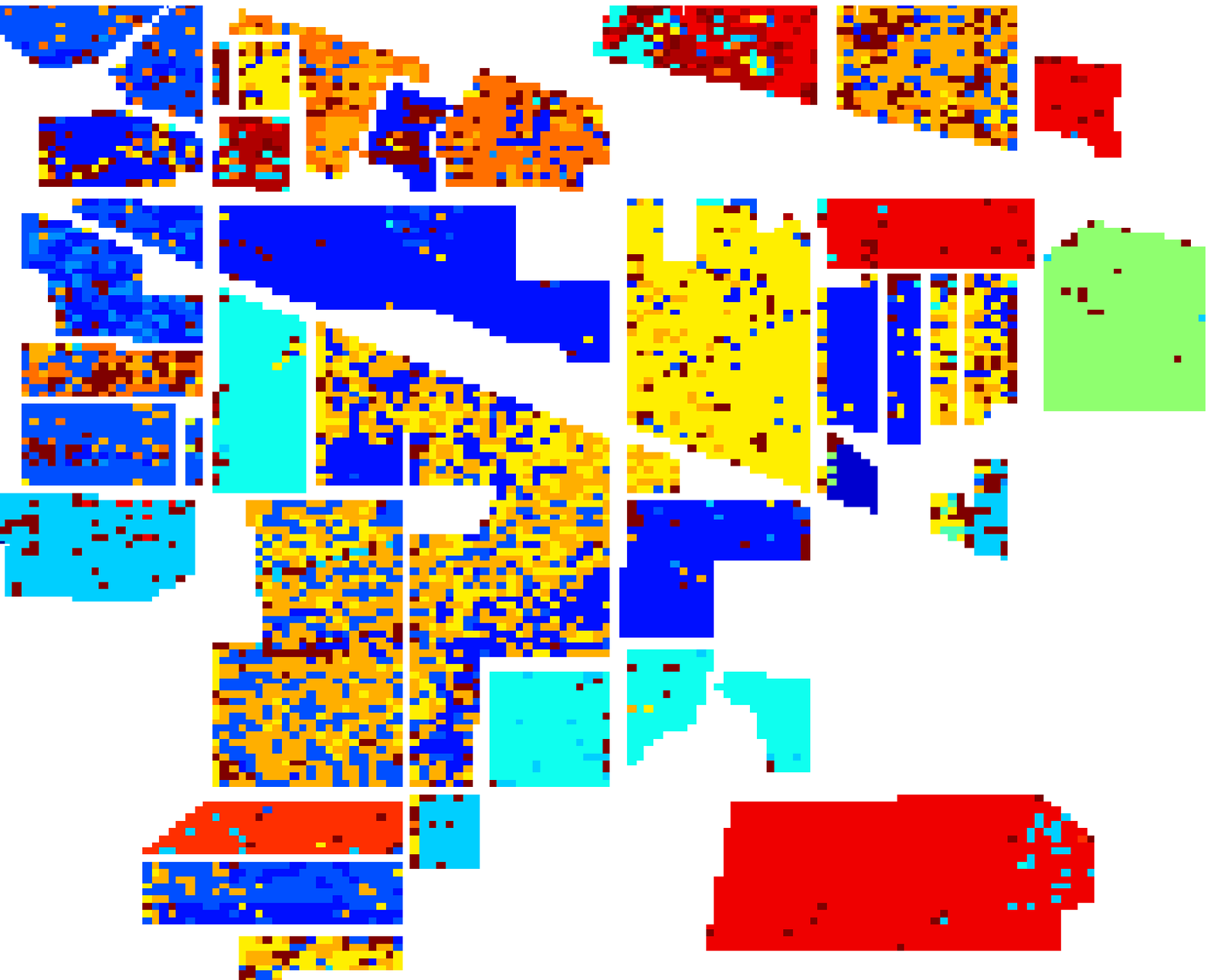}
\caption*{(d)}
\label{fig:figure2}
\end{minipage}
\hspace{0.2cm}
\begin{minipage}[b]{0.12\linewidth}
\centering
\includegraphics[width=\textwidth]{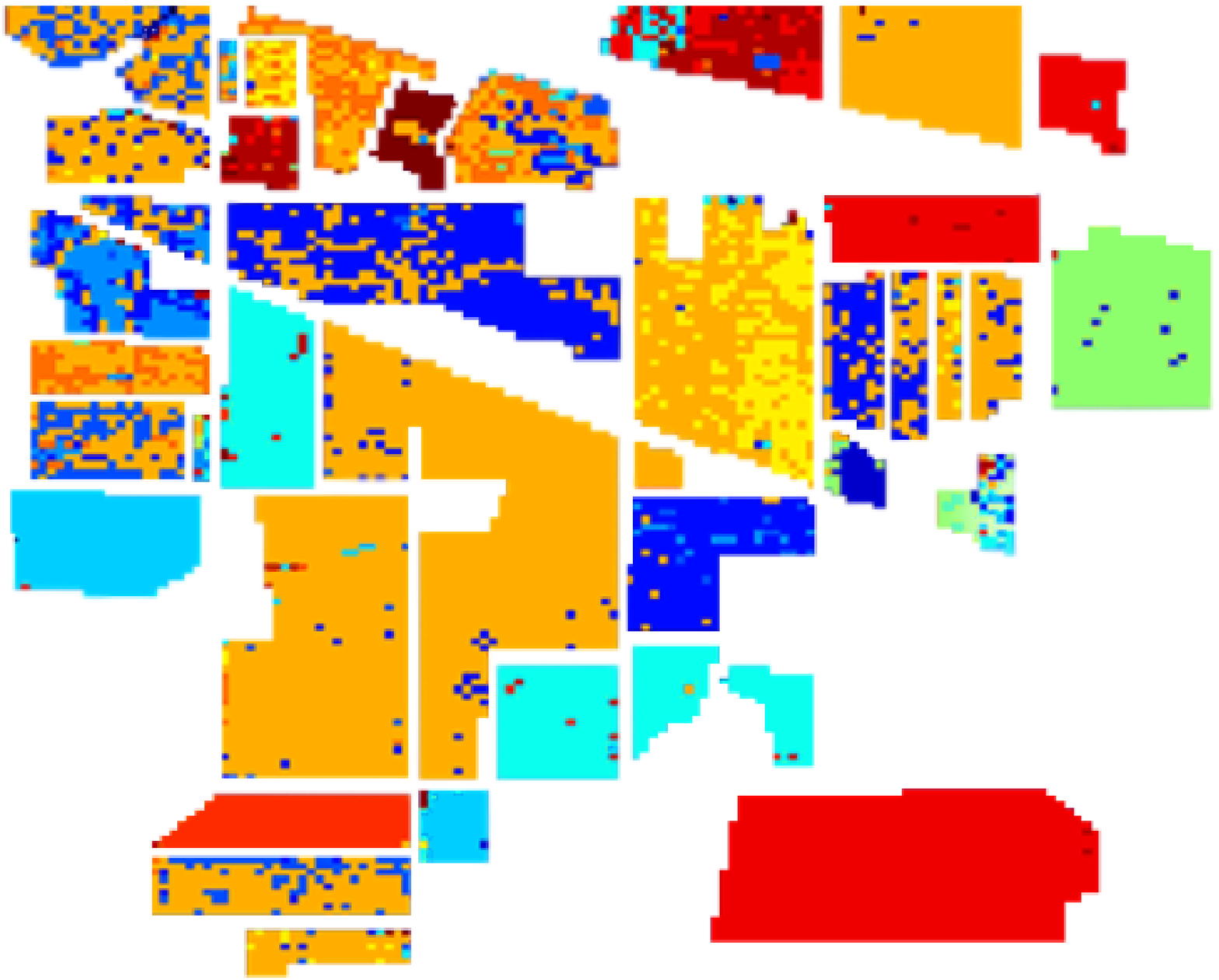}
\caption*{(e)}
\end{minipage}
\hspace{0.2cm}
\begin{minipage}[b]{0.12\linewidth}
\centering
\includegraphics[width=\textwidth]{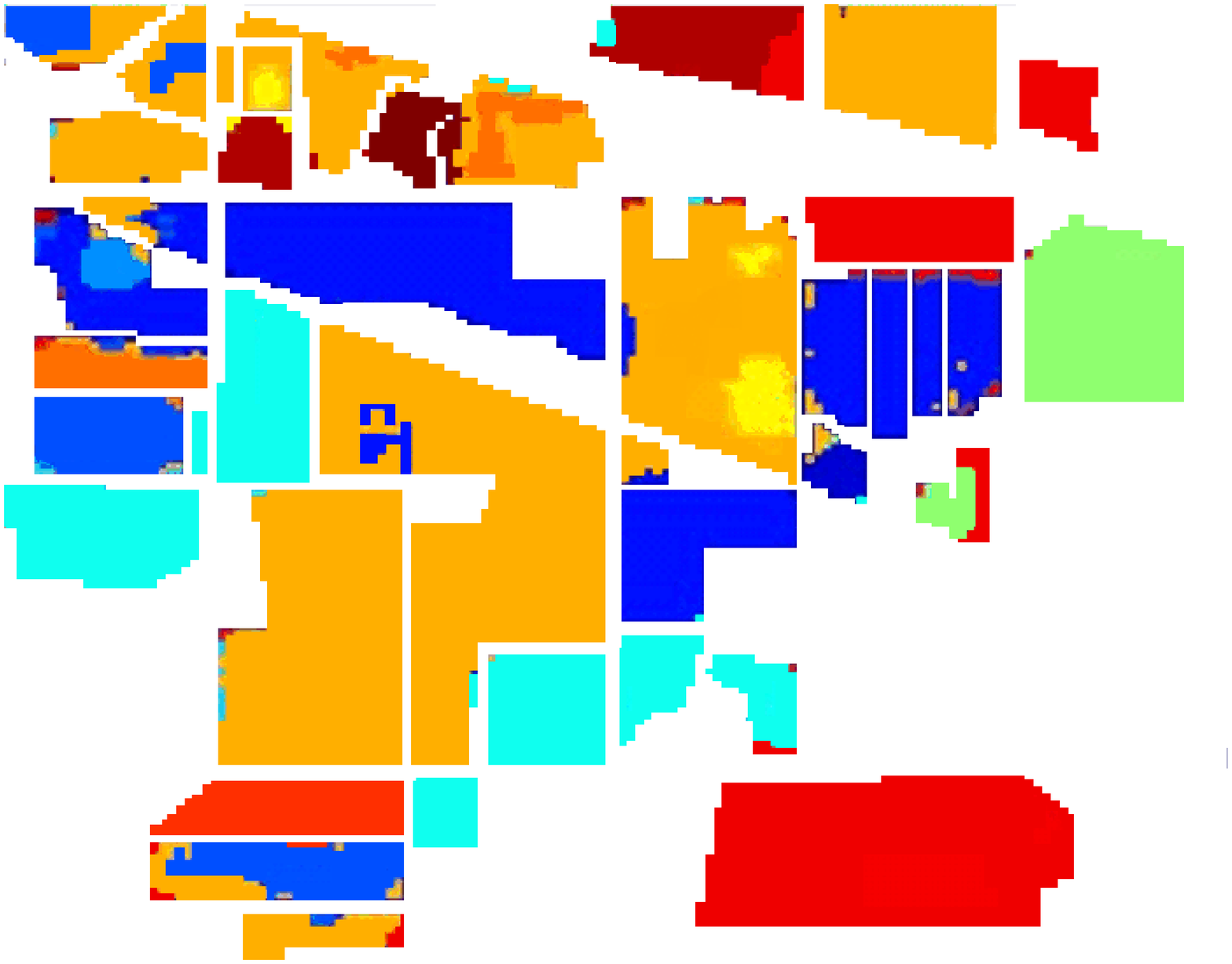}
\caption*{(f)}
\label{fig:figure2}
\end{minipage}
\\[0.2ex]
\begin{minipage}[b]{0.12\linewidth}\par\medskip
\centering
\includegraphics[width=\textwidth]{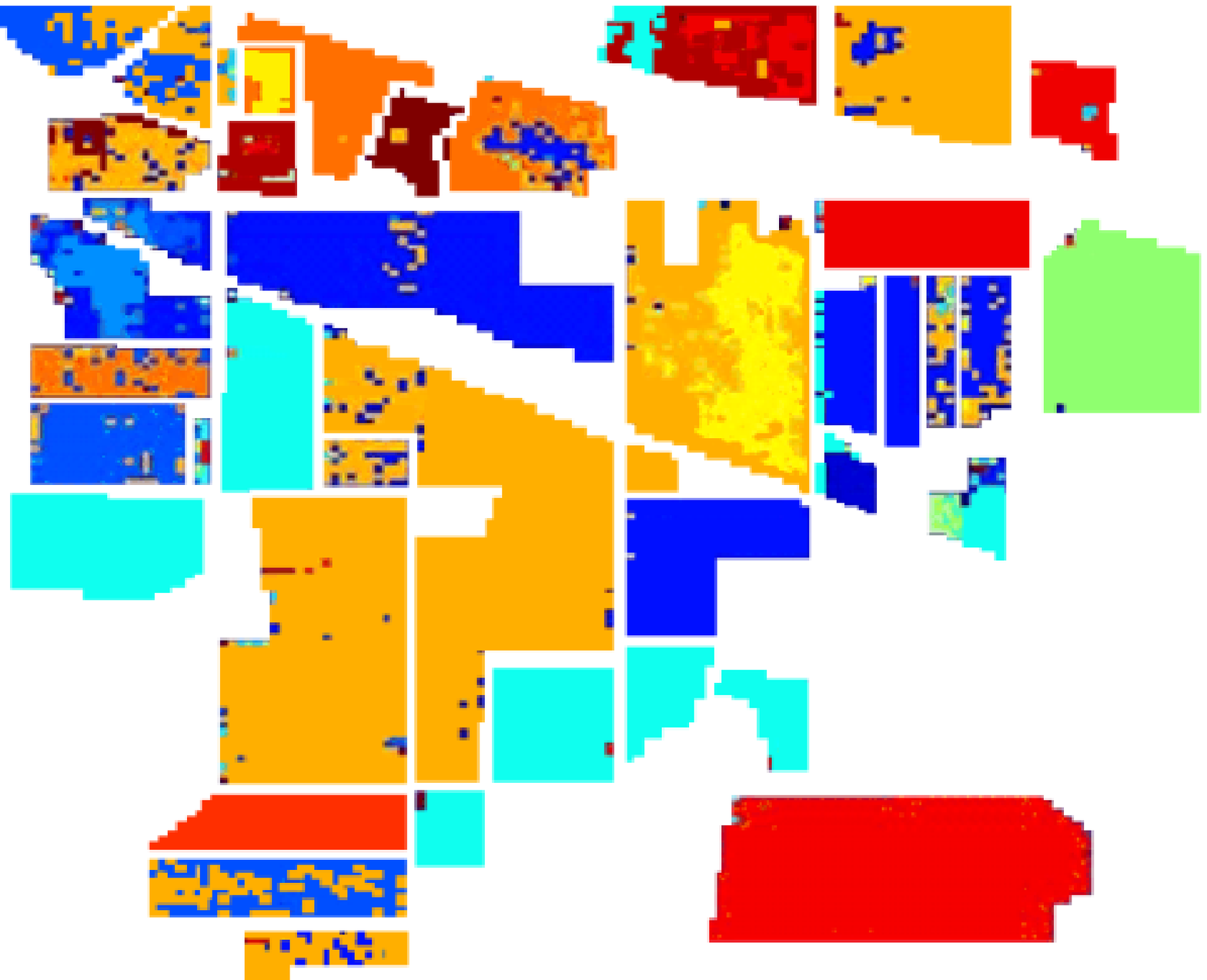}
\caption*{(g)}
\label{fig:figure1}
\end{minipage}
\hspace{0.2cm}
\begin{minipage}[b]{0.12\linewidth}\par\medskip
\centering
\includegraphics[width=\textwidth]{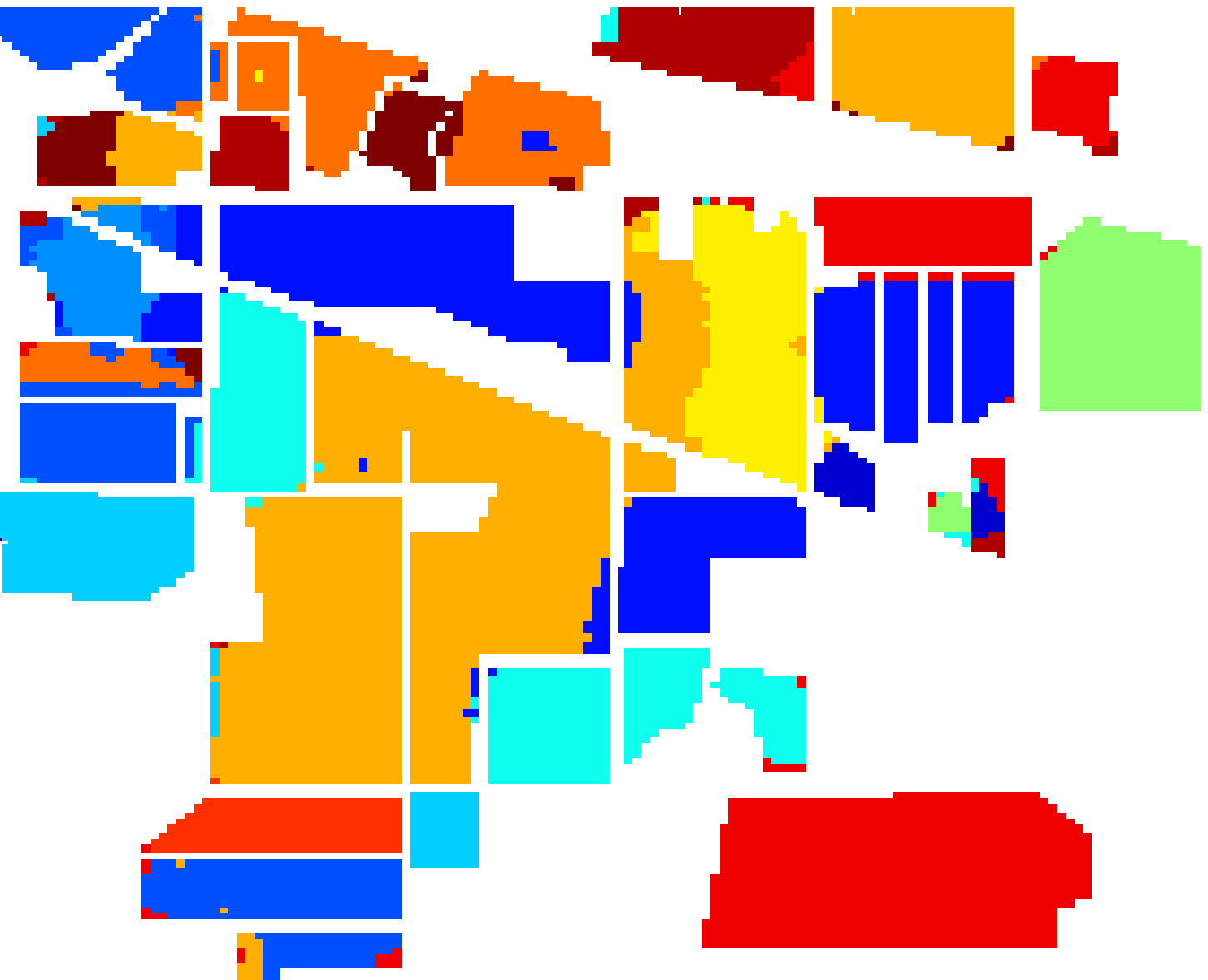}
\caption*{(h)}
\label{fig:figure1}
\end{minipage}
\hspace{0.2cm}
\begin{minipage}[b]{0.12\linewidth}\par\medskip
\centering
\includegraphics[width=\textwidth]{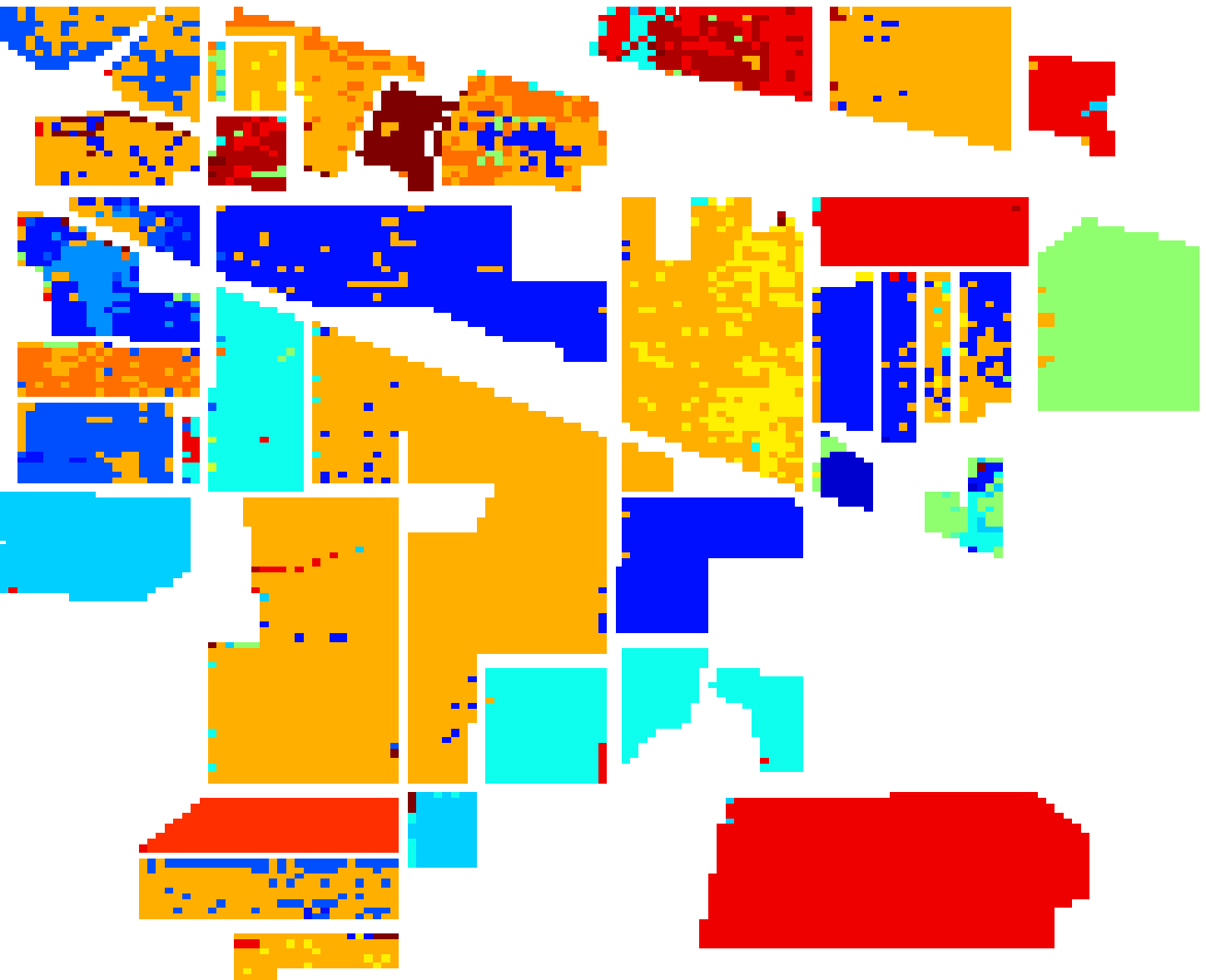}
\caption*{(i)}
\label{fig:figure1}
\end{minipage}
\hspace{0.2cm}
\begin{minipage}[b]{0.12\linewidth}\par\medskip
\centering
\includegraphics[width=\textwidth]{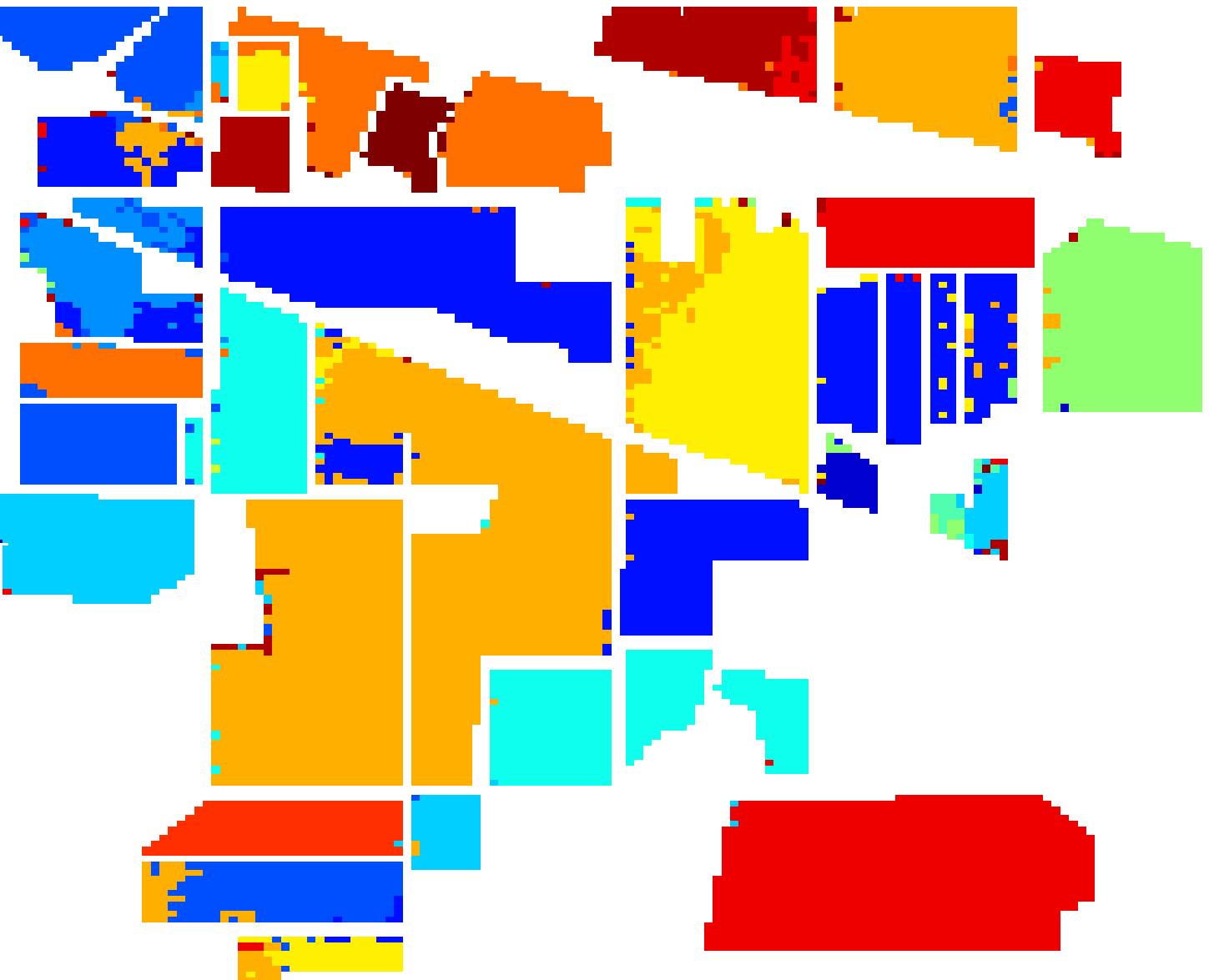}
\caption*{(j)}
\label{fig:figure1}
\end{minipage}
\hspace{0.2cm}
\begin{minipage}[b]{0.12\linewidth}
\centering
\includegraphics[width=\textwidth]{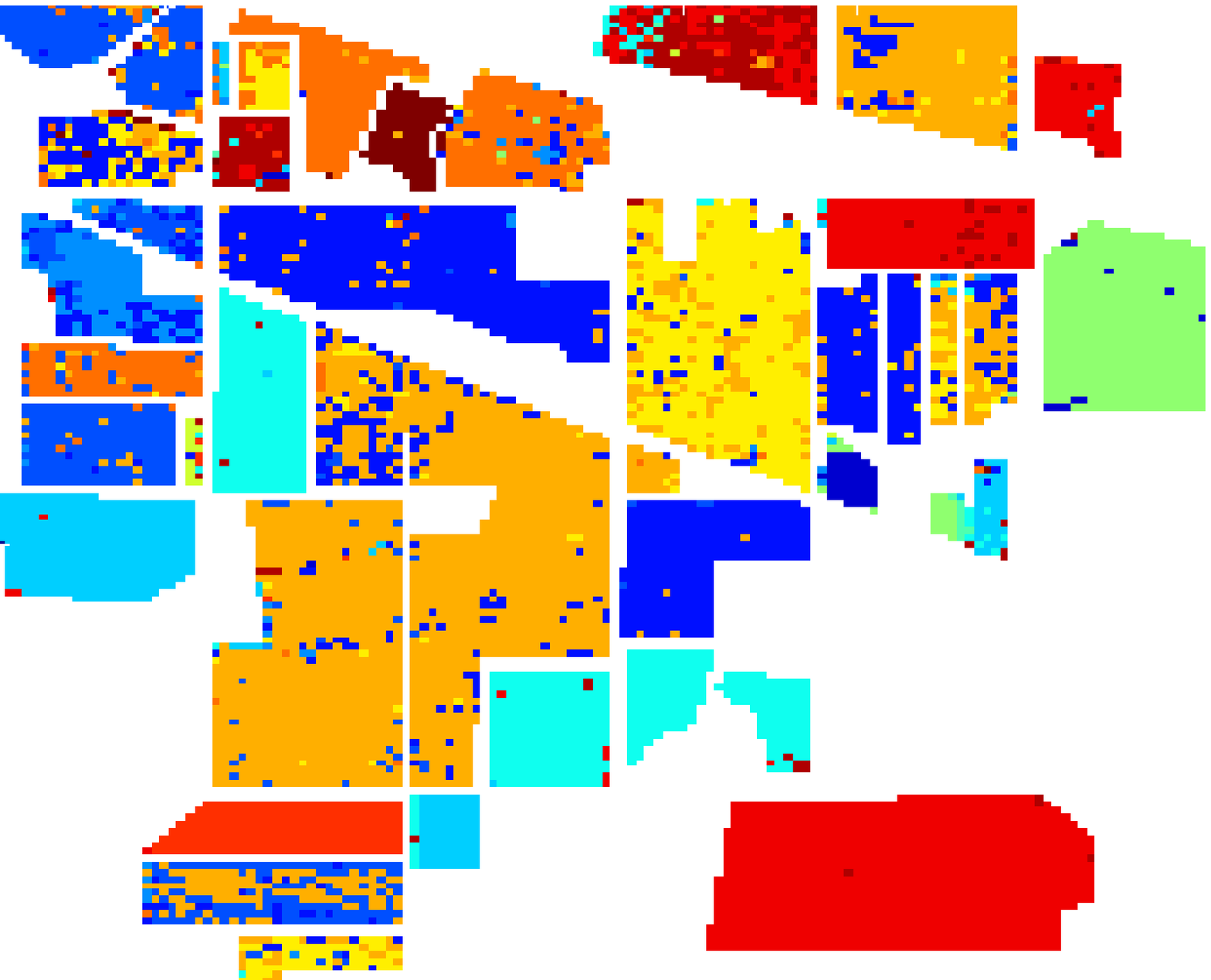}
\caption*{(k)}
\label{fig:figure1}
\end{minipage}
\hspace{0.2cm}
\begin{minipage}[b]{0.12\linewidth}
\centering
\includegraphics[width=\textwidth]{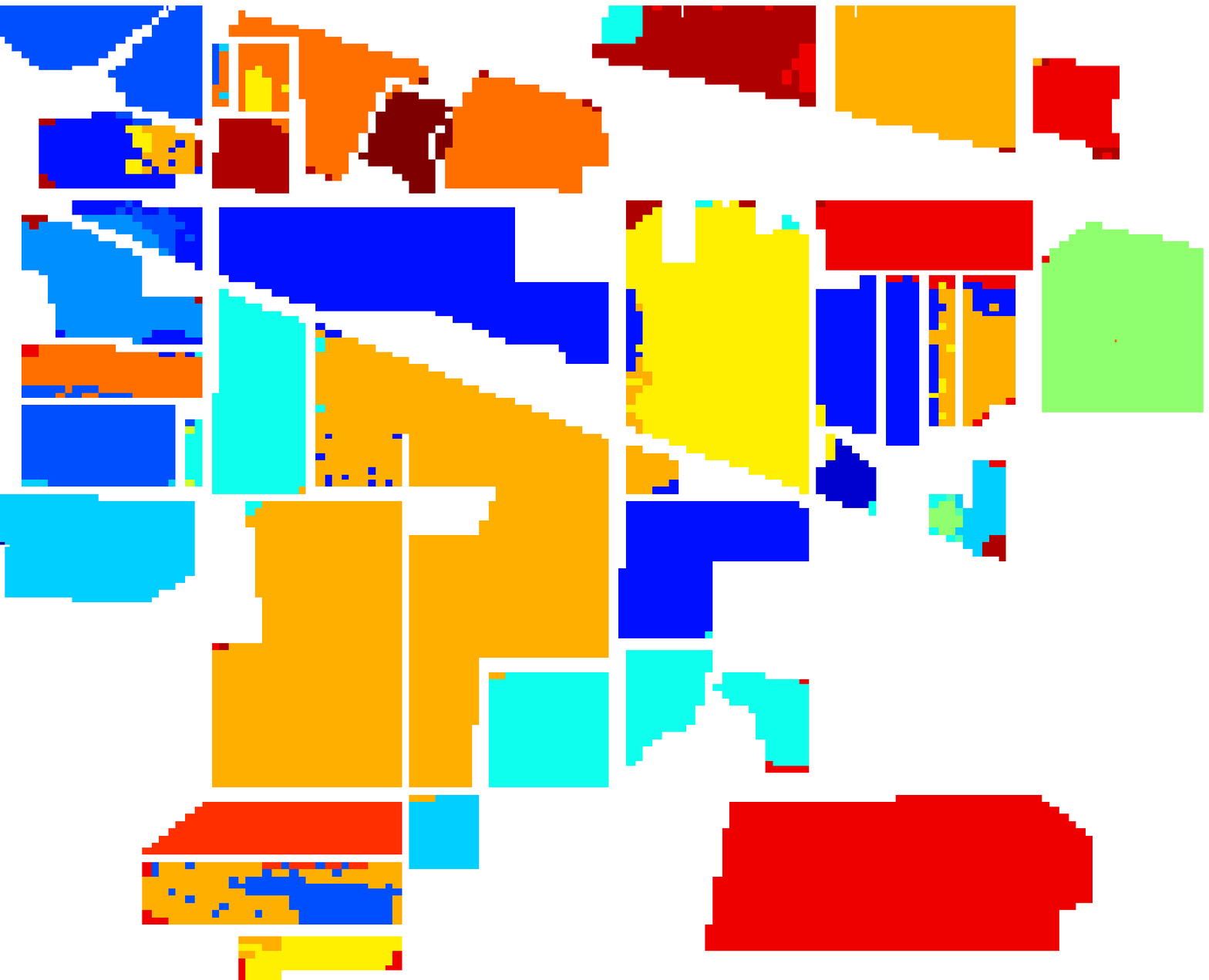}
\caption*{(l)}
\label{fig:figure1}
\end{minipage}
\caption{Results for the Indian Pine image: (a) ground truth, (b) training set and (c) test set. Classification map obtained by (d) SVM, (e) $\ell_1$-minimization using ADMM, (f) joint sparsity, (g) collaborative group sparsity, (h) collaborative sparse group sparsity, (i) low rank sparsity, (j) low rank group sparsity, (k) $\ell_1$ minimization via FSS and (l) Laplacian sparsity via FSS. }
\label{fig:indianpine}
\end{figure*}

%

To incorporate the structure of the dictionary, we now extend the low rank prior to group low rank prior, where the regularizer is obtained by summing up the rank of every group coefficient matrix,
\begin{equation}
\min_{\mathbf{X}}{\frac{1}{2}\lVert\mathbf{Y-AX}\rVert_F^2 + \lambda\sum\limits_{g\in G}w_g\lVert\mathbf{X}_g\rVert_*} .
\end{equation}
The low rank group prior is able to obtain the within-group sparsity by minimizing the nuclear norm of each group. Furthermore, the summation of nuclear norms empowers the proposed prior to obtain a group sparsity pattern. Hence, the low rank group prior is able to achieve sparsity both within and across groups by using only one regularization term.


%

\begin{table*}[ht]
  \scriptsize
    \begin{minipage}{.3\linewidth}
      \caption{Number of training and test samples for the Indian Pine image}
      \centering
    
        \begin{tabular}[ht]{| c || c | c | }
\hline
Class   & Train & Test \\
\hline
 1      &6 & 48 \\
 2   &137 &1297 \\
 3  &80& 754   \\
 4   &23 &211 \\
 5    &48&449 \\
 6   &72 & 675\\
 7   &3 & 23 \\
 8    &47 &442\\
 9   &2& 18 \\
10  &93 & 875 \\
11   &235 & 2233 \\
12    &59 &555\\
13   &21 &191 \\
14   &124&1170\\
15  &37 &343 \\
16   &10& 85 \\
\hline
{Total} &997 &9369\\
\hline
\end{tabular}
    \end{minipage}%
    \begin{minipage}{.7\linewidth}
      \centering
        \caption{Classification accuracy ($\%$) for the Indian Pine image using 997 ($10.64\%$) training samples}
\begin{tabular}[ht]{| c || c | c | c | c | c | c | c | c || c | c |  }
\hline
\multicolumn{2}{|c|}{Optimization Techniques} &\multicolumn{7}{c||}{ADMM/SpaRSA}&\multicolumn{2}{c|}{Feature Sign Search}\\
\hline
Class    & SVM &$\ell_1$  &JS &LS & GS & SGS & LR & LRG  & $\ell_1$ &LS\\
\hline
 1       &  77.08  &68.75   & 79.17  &85.42   & 79.17 & 87.50 & 75.00  & \textbf{91.67}& 66.67 &83.33 \\
 2      & 84.96 &58.84  & 81.94  &81.34  & 80.62 & 79.92 & 78.60 & 81.71  & 74.42   &\textbf{89.90}\\
 3   &  62.67   &24.40  & 56.67  &47.35  & 62.13 & 76.13 & 29.87  & \textbf{89.87} & 69.87 &78.38\\
 4  &  8.57   &49.52  & 27.62 &49.76  & 37.14 & 54.29 & 15.24  & 67.62  & 64.76 &\textbf{88.15}\\
 5     &  77.18  &81.88   & 85.46 &83.96  & 84.79 & 82.55 & 82.10  & 83.45  & 91.72 &\textbf{94.43}\\
 6    &  91.82   &96.88  & 98.36 &97.48  & \textbf{98.96} & 98.36 & 98.21  & 98.36  & 97.02 &98.52\\
 7    &  13.04   &0.00  & 0.00  &0.00   & 0.00  & 0.00 & 0.00    & 0.00  & \textbf{69.57} &0.00\\
 8    &  96.59  &96.59  & \textbf{100.00} &99.55 & 99.55 & 99.55 & 99.77  & 99.55  & 99.55  &\textbf{100.00}\\
 9   &  0.00  &5.56   & 0.00   &0.00   & 22.22 & 0.00 & 0.00   & 0.00 &\textbf{61.11} &0.00\\
10    &  71.30 &24.00    & 18.94 &31.89  & 39.95 & 45.58  &8.61   & 49.60  & 76.46 &\textbf{87.43}\\
11    &  35.25  &96.22   & 91.63 &94.58  & 91.99 & 93.02  & 97.12 & 92.35  & 87.62 &\textbf{98.84}\\
12    &  42.39  &32.97  & 45.29 &64.68  & 69.57 & 65.58  & 20.83 & 82.97  & 78.26  &\textbf{91.71}\\
13      &  91.05  &98.95   & 99.47 &99.48  & 99.47 & 98.95& 98.95   & 99.47  & 99.47  &\textbf{100.00}\\
14    &  94.85  &98.97   & 98.97 &99.49 & 98.80 & 99.31 & \textbf{99.83}    & 99.31   & 97.77 &99.57\\
15   &  30.70  &49.71  & 55.85 &63.84 & 50.58 & 80.99 &44.15    & \textbf{89.47}  & 53.80  &69.97\\
16    &  27.06  &88.24   & 95.29 &97.65 & 95.29 & \textbf{98.82} &97.65    & 97.65  & 85.88 &97.65\\
\hline
\textbf{OA}[$\%$]  & 64.94  &71.17   & 76.41 &79.40 & 80.19 & 83.19  &71.90   & 86.46  & 83.74 &\textbf{92.58}\\
\textbf{AA}[$\%$]  & 56.53  &60.72 &68.53   & 64.67 & 69.39 & 72.53  &59.14   & 76.43 & 79.62 &\textbf{79.87}\\
$\mathbf{\kappa}$  & 0.647  &0.695 & 0.737 &0.712 & 0.781 & 0.807 & 0.695           &0.843  & 0.833  &\textbf{0.923}\\
\hline

\end{tabular}
    \end{minipage} 
\end{table*}

\begin{table*}[ht]
\scriptsize
    \begin{minipage}{.3\linewidth}
      \caption{Number of training and test samples for the University of Pavia image}
      \centering
 \begin{tabular}{| c || c | c | }
\hline
Class   & Train &Test \\
\hline
 1    &139 &6713 \\
 2    &137 &1859 \\
 3   &100 &2107 \\
 4  &133 & 3303\\
 5   &68&1310 \\
 6   &135&4969 \\
 7   &95 & 1261 \\
 8   &131 &3747  \\
 9  &59 &967 \\
\hline
Total &997 &42926 \\
\hline
\end{tabular}
    \end{minipage}%
    \begin{minipage}{.7\linewidth}
      \centering
        \caption{Classification accuracy ($\%$) for the University of Pavia image using 997 ($2.32\%$) training samples}
\begin{tabular}{| c || c | c | c | c | c | c | c | c || c | c | }
\hline
\multicolumn{2}{|c}{Optimization Techniques} &\multicolumn{7}{|c||}{ADMM/SpaRSA}&\multicolumn{2}{c|}{Feature Sign Search}\\
\hline
Class   & SVM & $\ell_1$  &JS  &LS & GS & SGS & LR & LRG  &$\ell_1$ & LS\\
\hline
 1     &84.55   & 57.11    & 77.04  &95.08 & 94.01 & 97.90 & 91.16    &94.15  &72.14 &\textbf{95.85}\\
 2   &\textbf{82.45}   & 58.22  &67.98  & 66.70   & 70.04 & 68.04 & 69.73   &69.32  &59.62 &64.28\\
 3     &77.08   & 57.33    & 44.32 &77.55  & 79.45 & 73.56 & 75.80    &\textbf{79.73}  &66.21 &76.51\\
 4    &94.19   & 95.94    & 95.13   & 95.19 & 95.31 & 95.55    &95.94 &98.46 &97.67 &\textbf{98.97}\\
 5  &99.01   & \textbf{100.00}    & 99.85  &\textbf{100.00} & \textbf{100.00} & \textbf{100.00} & \textbf{100.00} &\textbf{100.00}  &99.85 &\textbf{100.00}\\
 6   &23.55   & 89.60    & 88.31 &96.60  & \textbf{100.00} & 99.74 & \textbf{100.00}  &99.96  &80.60 &98.63\\
 7    &2.06  & 83.27    & 84.38  &\textbf{96.59}  & 95.24  &95.56 & 95.06   &95.24  &86.76 &94.69\\
 8    &33.89  & 48,65    & 65.20 &67.36 & 62.24 & 44.84 & 65.24     &63.06  &75.95 &\textbf{95.76}\\
 9    &53.05  & 93.69    & \textbf{99.59}  &\textbf{99.59}   & 93.38 & 93.28 & 93.57  &94.00  &90.69 &98.35\\
\hline
\textbf{OA}[$\%$] & 69.84  & 66.51    & 74.05 &80.82 & 81.15 & 79.07  &80.81     &81.02  &71.41 &\textbf{81.84}\\
\textbf{AA}[$\%$] & 61.09  & 75.98    & 80.06 &88.80 & 87.73 & 85.36  &87.35     &87.93  &81.05 &\textbf{91.45}\\
$\mathbf{\kappa}$  &  0.569  & 0.628  & 0.681  &0.758  & 0.675  &  0.624 & 0.611  & 0.66 &0.672 &\textbf{0.781}\\
\hline
\end{tabular}
    \end{minipage} 
\end{table*}

\section{Results and Discussion}
\label{sec:majhead}
\subsection{Datasets}
We evaluate various structured sparsity priors on two different hyperspectral images and one toy example. The first hyperspectral image to be assessed  is the Indian Pine, acquired by Airborne Visible/Infrared Imaging Spectrometer (AVIRIS), which generates 220 bands, of which 20 noisy bands are removed before classification. The spatial dimension of this image is $145\times 145$, which contains 16 ground-truth classes, as shown in Table \uppercase\expandafter{\romannumeral1}. We randomly choose 997 pixels ($10.64\%$ of all the labelled pixels) for constructing the dictionary and use the remaining pixels for testing. The second image is the University of Pavia, which is an urban image acquired by the Reflective Optics System Imaging Spectrometer (ROSIS), contains $610\times340$ pixels. It generates 115 spectral bands, of which 12 noisy bands are removed. There are nine ground-truth classes of interests. For this image, we choose 997 pixels ($2.32\%$ of all the labelled pixels) for constructing the dictionary and the remaining pixels for testing, as shown in Table \uppercase\expandafter{\romannumeral3}.  The toy example consists of two different classes (class 2 and 14 of the Indian Pine test set), and each class contains 30 pixels. The dictionary is the same as that for the Indian Pine. The toy example is used to evaluate the various sparsity patterns generated by the  different structured priors.

\subsection{Models and Methods}
 The tested structured sparse priors are: ({\it \romannumeral1}) joint sparsity (JS), ({\it \romannumeral2}) Laplacian sparsity (LS), ({\it \romannumeral3}) collaborative group sparsity (GS), ({\it \romannumeral4}) sparse group sparsity (SGS), ({\it \romannumeral5}) low rank prior (LR) and ({\it \romannumeral6}) low rank group prior (LRG), corresponding to Eqs. (7), (10), (12), (14), (16) and (17), respectively. For SRC, the parameters $\lambda, \lambda_1$ and $\lambda_2$ of different structured priors range from $10^{-3}$ to $0.1$. Performance on the toy example will be visually examined by the difference between the desired sparsity regions and the recovered ones. For the two hyperspectral images, classification performance is evaluated by the overall accuracy (OA), average accuracy (AA), and the $\kappa$ coefficient measure on the test set. For each structured prior, we present the result with the highest overall accuracy using cross validation.  A linear SVM is implemented for comparison, whose parameters are set in the same fashion as in \cite{Chen}. 

In experiments, joint sparsity, group sparsity and low rank priors are solved by ADMM \cite{Boyd}, while CHiLasso  and Laplacian prior  are solved by combining SpaRSA \cite{Wright} and ADMM.  In addition, in conformity with previous work \cite{Gao}, the Laplacian regularized Lasso is also solved by a modified feature sign search (FSS) method.  In this paper, we try to present a fair comparison among all priors. According to the optimization technique, we sort the structured priors into two categories: ({\it \romannumeral1}) priors solved by  ADMM and SpaRSA  and ({\it \romannumeral2}) priors solved by FSS-based method. The first row of Table \uppercase\expandafter{\romannumeral2} and Table \uppercase\expandafter{\romannumeral4} show the methods used to implement the sparse recovery for each structured prior. 

\begin{table}[ht]
        \caption{Computation time (in seconds) for the Indian Pine image}
        \centering
\begin{tabular}[ht]{| c | c | c | c | c | c | c | c | c | c | c |  }
\hline
\multicolumn{7}{|c}{ADMM/SpaRSA}&\multicolumn{2}{|c|}{FFS}\\
\hline
$\ell_1$  &JS &LS  & GS & SGS & LR & LRG &LS & $\ell_1$ \\
\hline
1124   & 1874  &4015 & 2811 & 2649 & 4403  & 2904  & 1124 &11628 \\
\hline

\end{tabular}
\label{table:time}
\end{table}

\subsection{Results}

Sparsity patterns of the toy example are shown in Fig. \ref{fig:sp}. The expected sparsity regions are shown in Fig. \ref{fig:sp}(a), where the y-axis labels the dictionary atom index and x-axis labels the test pixel index. The red and green regions correspond to the ideal locations of the active atoms for the class 2 and 14, respectively. Nonzero coefficients that belong to other classes are shown in blue dots. The joint sparsity, Fig. \ref{fig:sp} (c), shows clear row sparsity pattern, but many rows are mistakenly activated. As expected, active atoms in Fig. \ref{fig:sp} (d), (e) and (g) demonstrate group sparsity patterns. Comparing the GS (d) and SGS (e), it is observed that most of the atoms are deactivated within groups  using SGS. The low rank group prior (g) demonstrates a similar sparsity pattern as that of SGS. For the Laplacian sparsity (h), similarity of sparse coefficients that belong to the same classes is clearly visible. 

 Table \uppercase\expandafter{\romannumeral2} and Fig. \ref{fig:indianpine} show the performance of SRCs with different priors on the Indian Pine image. A spatial window of $9\times9$ ($T=81$) is used since this image consists of mostly large homogeneous regions. Among SRCs with different priors, the worst result occurs when we use simple $\ell_1$-ADMM. Joint sparsity prior gives better result than the low rank prior. This is due to the large areas of homogeneous regions in this image, which favors the joint sparsity model. The highest OA is given by the Laplacian sparsity prior  via FFS, such a high performance is partly contributed to the accurate sparse recovery of the feature sign search method. Both SGS and LRG outperform GS. We can see that among ADMM-based based methods, the low rank group prior yields the smoothest result.  The computational time of various structured priors for Indian Pine image are shown in Table \ref{table:time}. Among ADMM/SpaRSA-based methods, LRG, GS and SGS  take roughly similar time ($\sim$2500s) to process the image, while LR and JS require longer time ($\sim$4000s). LS via FFS significantly impedes the computational efficiency.

Results for the University of Pavia image are shown in Table \uppercase\expandafter{\romannumeral4}.  The window size for this image is $5\times5 $ ($T=25$) since many narrow regions are present in this image. The group sparsity prior gives the highest OA among the priors optimized by ADMM. The low rank sparsity prior gives a much better result than joint sparsity since this image contains many small homogeneous regions.  The Laplacian sparsity prior  via FFS gives the highest OA performance. However, the difference between performance of various structured priors is quite small.

\section{CONCLUSION}
\label{sec:page}
This paper reviews five different structured sparse priors and proposes a low rank group sparsity prior.  Using these structured priors, classification results of SRCs on HSI are generally improved  when  compared with the classical $\ell_1$ sparsity prior. The results have confirmed that the low rank prior is a more flexible constraint compared with the joint sparsity prior, while the latter works better on large homogeneous regions. Imposing the group structured prior on the dictionary always gives higher overall accuracy compared with the $\ell_1$ prior. We have also observed that the performance is not only determined by the structured priors, but also depend on the corresponding optimization techniques.


\bibliographystyle{IEEEtran}


\begin{thebibliography}{99}

{\footnotesize
\bibitem{Plaza}
A. Plaza, J.  Benediktsson, J.  Boardman, J.  Brazile, L. Bruzzone, G. Camps-Valls, J. Chanussot, M. Fauvel, P. Gamba, A. Gualtieri, M.  Marconcini, J. Tiltoni and G. Trianni, ``Recent advances in techniques for hyperspectral image processing,'' {\it Remote  Sens.  Envir.}, vol. 113, no. s1, pp. s110-s122, Sept. 2009.



\bibitem{GCampsValls}
G. Camps-Valls, L. Gomez-Chova, J. Mu\~{n}oz-Mar\`{i}, J. Vila-Franc\'{e}s and J. Calpe-Maravilla, ``Composite kernels for hyperspectral image classification,'' {\it IEEE Geosci. Remote Sens. Lett.}, vol. 3, no. 1, pp. 93-97, Jan. 2006.

\bibitem{Chova}
L. Gómez-Chova, G. Camps-Valls, J. Muñoz-Marí and J. Calpe-Maravilla, ``Semi-supervised image classiﬁcation with Laplacian support vector machines,'' {\it IEEE Geosci. Remote Sens. Lett.}, vol. 5, no. 3, pp. 336-340, Jul. 2008.

\bibitem{JZhu}
J. Zhu , S. Rosset , T. Hastie and R. Tibshirani, ``1-norm support vector machines,'' {\it NIPS}, vol. 16, pp. 16-23, Dec. 2003.

\bibitem{JWright2}
J. Wright, A. Yang, A. Ganesh, S. Sastry and Y. Ma, ``Robust face recognition via sparse representation,'' {\it IEEE Trans. Pattern Anal. Mach. Intell.}, vol. 31, no. 2, pp. 210-227, Feb. 2009.

\bibitem{Marial}
J. Wright, J. Mairal, G. Sapiro, T.S. Huang, S. Yan, ``Sparse Representation for computer vision and pattern recognition,'' {\it Proceed. IEEE}, vol. 98, no. 6,  pp. 1031-1044, Apr. 2010.


\bibitem{Chen}
Y. Chen, M. Nasrabadi and T. Tran, ``Hyperspectral image classification using dictionary-based sparse representation,'' {\it IEEE Trans. Geosci. Remote Sens.}, vol. 49, no. 6, pp. 2287-2302, Oct. 2011.

\bibitem{Haq}
Q. Haq, L. Tao, F. Sun and S. Yang, ``A fast and robust sparse approach for hyperspectral data classification using a few labeled samples,'' {\it IEEE Trans. Geosci. Remote Sens.}, vol. 50, no. 10, pp. 3973-3985, June 2012.

\bibitem{RJi}
R. Ji,  Y. Gao, R. Hong, Q.  Liu, D. Tao and X. Li, ``Spectral-spatial constraint hyperspectral image classification,'' {\it  IEEE Trans. Geosci. Remote Sens.}, vol. PP,  no. 99, pp. 1-13, June 2013.



\bibitem{MIordache}
M. Iordache, J. Bioucas-Dias and A. Plaza, ``Sparse unmixing of hyperspectral data,'' {\it IEEE Geosci. Remote Sens.,}, vol. 49, no. 6, pp. 2014-2039, June 2011.




\bibitem{Tropp}
J. Tropp, A. Gilbert and M. Strauss, ``Algorithms for simultaneous sparse approximation. Part I: Greedy pursuit,'' {\it Signal Processing}, vol. 54, no. 12, pp. 4634-4643, Dec. 2006.


\bibitem{Ewout}
E. Berg and M. Friedlander, ``Joint-sparse recovery from multiple measurements,'' {\it IEEE Trans. Information Theory.}, vol. 56, no. 5, pp. 2516-2527, Apr. 2010.

\bibitem{Gao}
S. Gao, I. Tsang and L. Chia, ``Laplacian sparse coding, hypergraph Laplacian sparse coding, and applications,'' {\it IEEE Trans. Pattern Anal. Mach. Intell.}, vol. 35, no. 1, pp. 92-104, Jan. 2013.

\bibitem{Liu}
G. Liu, Z. Lin, S. Yan, J. Sun, Y. Yu and Y. Ma, ``Robust recovery of subspace structures by low-rank representation,'' {\it IEEE Trans. Pattern Anal. Mach. Intell.}, vol. 35, no. 1, pp. 171-184, Jan. 2013.

\bibitem{Rakotomamonjy}
A. Rakotomamonjy, ``Surveying and comparing simultaneous sparse approximation (or group-lasso) algorithms,'' {\it  Signal Processing}, vol. 91, no. 7, pp. 1505-1526, July 2011.

\bibitem{Kim}
S.Kim and E. Xing, ``Tree-guided group lasso for multi-task regression with structured sparsity,'' {\it ICML}, vol. 6, no. 3, pp. 1095-1117, June 2010.

\bibitem{Sprechmann}
P. Sprechmann, I. Ramirez, G. Sapiro and Y. Eldar, ``C-HiLasso: a collaborative hierarchical sparse modeling framework,'' {\it IEEE Trans. Signal Processing}, vol. 59, no. 9, pp. 4183-4198, Oct. 2011.


\bibitem{Qian}
Y. Qian, M. Ye and J. Zhou, ``Hyperspectral image classification based on structured sparse logistic regression and three-dimensional wavelet texture features,'' {\it IEEE Trans. Geosci. Remote Sens.}, vol. 51, no. 4, pp. 2276-2291, Apr. 2012.


\bibitem{Elhamifar}
E. Elhamifar and R. Vidal, ``Sparse subspace clustering,'' {\it CVPR}, pp. 2790-2797, June 2009.


\bibitem{Boyd}
S. Boyd, N. Parikh, E. Chu, B. Peleato and J. Eckstein, ``Distributed optimization and statistical learning via the alternating direction method of multipliers,'' {\it FTML.}, vol. 3, no. 1, pp. 1-122, Jan. 2010.


\bibitem{Wright}
S. Wright, R. Nowak and M. Figueiredo, ``Sparse reconstruction by separable approximation,'' {\it IEEE Trans. Signal Processing}, vol. 57, no. 7, pp. 2479-2493, July 2009.

}
\end{thebibliography}



\end{document}